\documentclass[lettersize,journal]{IEEEtran}
\usepackage{amsmath,amsfonts}
\usepackage{array}
\usepackage[caption=false,font=normalsize,labelfont=sf,textfont=sf]{subfig}
\usepackage{textcomp}
\usepackage{stfloats}
\usepackage{url}
\usepackage{verbatim}
\usepackage{graphicx}
\usepackage{cite}
\usepackage{adjustbox}
\usepackage{xcolor}
\usepackage{pifont}
\newcommand{\cmark}{\ding{51}} 
\newcommand{\xmark}{\ding{55}}
\usepackage[ruled, lined, linesnumbered, commentsnumbered, longend]{algorithm2e}
\usepackage[colorlinks, linkcolor=green,
anchorcolor=green,
citecolor=green]{hyperref}
\begin{document}

\title{AO2-DETR: Arbitrary-Oriented Object Detection Transformer}
\author{Linhui Dai, Hong Liu$^*$ \IEEEmembership{Member, IEEE}, Hao Tang, Zhiwei Wu, Pinhao Song
\thanks{Linhui Dai, Hong Liu, and Pinhao Song are with the Key Laboratory of Machine Perception, Shenzhen Graduate School, Peking University, Beijing 100871, China (Email: dailinhui@pku.edu.cn, hongliu@pku.edu.cn, pinhaosong@pku.edu.cn).} 
\thanks{Hao Tang is with the Computer Vision Lab, ETH Zurich, Zurich, Switzerland (Email: hao.tang@vision.ee.ethz.ch).} 
\thanks{Zhiwei Wu is with the School of Software Engineering, South China University of Technology, Guangzhou, Guangdong, China (Email: zhiwei.w@qq.com).} 
}


\markboth{Journal of \LaTeX\ Class Files,~Vol.~14, No.~8, May~2022}%
{Shell \MakeLowercase{\textit{et al.}}: A Sample Article Using IEEEtran.cls for IEEE Journals}


\maketitle

\begin{abstract}
  Arbitrary-oriented object detection (AOOD) is a challenging task to detect objects in the wild with arbitrary orientations and cluttered arrangements. Existing approaches are mainly based on anchor-based boxes or dense points, which rely on complicated hand-designed processing steps and inductive bias, such as anchor generation, transformation, and non-maximum suppression reasoning. Recently, the emerging transformer-based approaches view object detection as a direct set prediction problem that effectively removes the need for hand-designed components and inductive biases. In this paper, we propose an Arbitrary-Oriented Object DEtection TRansformer framework, termed AO2-DETR, which comprises three dedicated components. More precisely, an oriented proposal generation mechanism is proposed to explicitly generate oriented proposals, which provides better positional priors for pooling features to modulate the cross-attention in the transformer decoder. An adaptive oriented proposal refinement module is introduced to extract rotation-invariant region features and eliminate the misalignment between region features and objects. And a rotation-aware set matching loss is used to ensure the one-to-one matching process for direct set prediction without duplicate predictions.
  Our method considerably simplifies the overall pipeline and presents a new AOOD paradigm. Comprehensive experiments on several challenging datasets show that our method achieves superior performance on the AOOD task.     
\end{abstract}

\begin{IEEEkeywords}
Oriented Object Detection, Detection Transformer, Oriented Proposals, Feature Refinement.
\end{IEEEkeywords}
\section{Introduction}
\begin{figure*}
  \centering
    \subfloat[Two Stage: RoI Transformer \cite{roitrans}]{\includegraphics[width=0.3\linewidth, keepaspectratio]{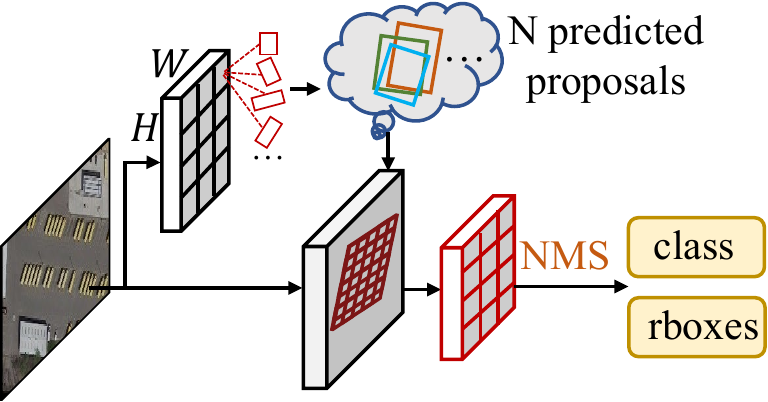}\label{fig1a}}
    \subfloat[Single Stage: R$^3$Det \cite{R3Det}]{\includegraphics[width=0.32\linewidth, keepaspectratio]{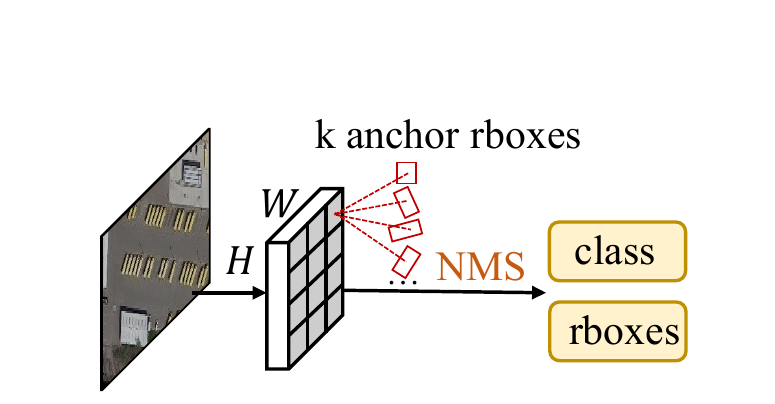}\label{fig1b}}
    \subfloat[Transformer-based: AO2-DETR]{\includegraphics[width=0.32\linewidth, keepaspectratio]{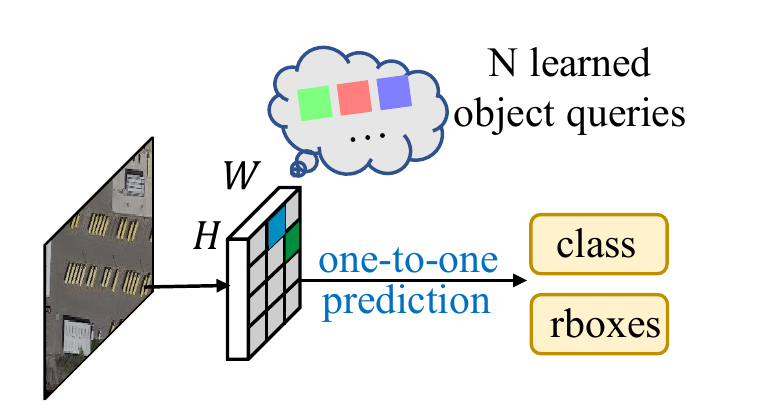}\label{fig1c}}
     \caption{Comparisons of different oriented object detection pipelines. (a) In two-stage detectors, a small set of $N$ candidates are selected from dense object candidates by rotated region proposal networks (RRPN), and then extract image features with corresponding regions by rotated roi pooling operation, e.g. RoI Transformer \cite{roitrans}. (b) In single-stage detectors, $H \times W \times k$ candidates enumerate on all image grids, e.g. R$^3$Det\cite{R3Det}. (c) The proposed transformer-based oriented object detector: AO2-DETR, directly outputs the predicted oriented boxes, without prior boxes or complex pre/post-processing steps.}
     \label{fig:intro}
  \end{figure*}

\IEEEPARstart{A}{rbitrary-oriented} object detection (AOOD) is a recently-emerged challenging problem in computer vision, which plays an important role in the field of aerial images \cite{ReDet,beyond}, smart retail \cite{drn}, and scene text \cite{rrpn}. Unlike generic object detection in nature images, oriented object detection has fundamental difficulties, including often distributed with arbitrary orientation, densely packed, or has highly complex backgrounds. 
Many recent oriented detection models have employed convolutional neural networks (CNNs) to achieve promising results. These methods could be roughly categorized into two types: anchor-based methods \cite{s2anet, scrdet, R3Det, ReDet} and anchor-free methods \cite{beyond, DARDet}. The anchor-based methods need to design the size and preset angle of the anchors manually. For instance, Ma et al. \cite{rrpn} introduce a rotation region proposal network to generate rotated proposals, which places 54 anchors with different angles, scales and aspect ratios. However, abundant anchors cause redundant computation and memory load. To address this issue, RoI Transformer \cite{roitrans} learns spatial transformations from horizontal Region of Interests (HRoIs) to rotated RoIs (RRoIs), as shown in Fig. \ref{fig1a}. Oriented R-CNN \cite{orientedrcnn} generates oriented proposals by directly learning midpoint offset representation. Meanwhile, R$^3$Det \cite{R3Det} as a one-stage approach generates oriented proposals directly and uses a feature refinement module to realize feature reconstruction and alignment, as shown in Fig. \ref{fig1b}. Nevertheless, these methods still require manual preset boxes and complex hyperparameters to achieve promising results. Therefore, several anchor-free methods \cite{fcosr,ienet,DARDet,DAFNe, beyond} are proposed in the AOOD task. For example, CFA \cite{beyond} models the object layout as a convex-hull, then refines the predicted convex-hull and makes it adapt to densely packed objects, which consists of two stages: convex-hull generation and adaptation. The anchor-free methods directly treat grid points in the feature map as object candidates and largely simplify the detection pipeline.

However, both rotated box candidates and point candidates have a common problem: each object will produce redundant and approximately duplicate predictions. In addition, it is necessary to carry out complicated hand-designed processing steps and inductive biases, e.g., anchor generation, anchor transformation, and non-maximum suppression (NMS) reasoning. Recently, transformer-based detectors \cite{transformer,detr,deformable,anchordetr} have been promoted as being dynamic, attentive, and can directly output predictions without complicated hand-designed processing steps and inductive biases. Set-to-set encoder-decoder models have emerged as a competitive way to model generic object detection. The self-attention and cross-attention operations in transformers are designed to be permutation-invariant. They can enable adaptive receptive fields for oriented objects, making them a natural candidate for processing rotated and irregularly placed objects. Motivated by this observation, we ask the following question: can we leverage transformers to learn an oriented object detector without relying on hand-designed inductive biases? 

Intuitively, directly extending the transformer-based detectors into the AOOD task by only adding additional angle prediction values will suffer from several issues: (1) Misalignment: the learned positional embeddings, which are called object queries, are horizontal. When object queries in the transformer decoder are decoded with the feed-forward network (FFN), it will typically lead to misalignment between region features and rotated objects. (2) Cluttered features: due to the various orientation and dense distribution of objects in aerial images, the learned horizontal proposals (object queries) may contain more background areas or multiple objects, resulting in cluttered extracted features. (3) Limited matching: the regular operations in transformer-based detectors have limited generalization to rotation and scale variations. Due to the highly diverse directions of objects, it is often intractable to acquire accurate matching with all objects by using object queries with the horizontal direction.

In this paper, we aim to alleviate the above issues for the challenging transformer-based oriented detection problem. We propose a transformer-based arbitrary-oriented object detection framework called AO2-DETR (see Fig. \ref{fig1c}). The proposed framework has three dedicated components to address oriented object detection settings: an oriented proposal generation (OPG) mechanism, an adaptive oriented proposal refinement (OPR) module, and a rotation-aware set matching loss. Specifically, \textbf{for the problem of misalignment and cluttered features}, we explore a novel oriented proposal generation mechanism for generating oriented region proposals as object queries, which are fed into the decoder as initial rotated boxes for adaptive oriented proposal refinement. The oriented proposals present a novel query formulation and provide a better positional prior for pooling features by predicting the orientation of each bounding box in addition to center and size. Next, we present the adaptive oriented proposal refinement module to alleviate the misalignment between the axis-aligned features and the arbitrary oriented objects. The initial position information is encoded by feature interpolation, and then the learned oriented proposals are adaptively adjusted by the refinement convolutional network. Consequently, \textbf{for the problem of limited matching}, we introduce a rotation-aware set matching loss, which allows AO2-DETR to infer the oriented bounding boxes directly without prior boxes or complex pre/post-processing steps. 

In conclusion, the main contributions of this paper can be summarized as follows:
\begin{itemize}
    \item We propose a transformer-based arbitrary-oriented object detector AO2-DETR, which eliminates the need for multiple anchors and complex pre/post-processing. We hope that our method can open up the possibility of developing various paradigms for the AOOD task.

    \item We design an oriented proposal generation (OPG) mechanism, which guides the network to generate oriented proposals with direction information. The oriented proposals can be used to solve the misalignment problem and provide a better positional prior for pooling features to modulate the cross-attention.
    
    \item We introduce a novel adaptive oriented proposal refinement (OPR) module into the transformer architecture. The OPR module dynamically adjusts the oriented proposals according to the learned context information by a feature alignment module and a larger receptive field, which can significantly reduce the gap between the oriented proposals and the ground-truth. In addition, we add a rotation-aware set matching loss in the one-to-one matching process to ensure the correct match between the predicted boxes and ground truth.
    \item The extensive experiments on four public datasets: DOTA-v1.0 \cite{dota}, DOTA-v1.5 \cite{dota}, SKU110K-R \cite{sku110k}, and HRSC2016 \cite{hrsc2016} demonstrate the effectiveness of the proposed model. AO2-DETR achieves the state-of-the-art performance among anchor-free and single-stage methods on these four datasets. Our code will be released at \url{https://github.com/Ixiaohuihuihui}.
\end{itemize}

\section{Related Works}

\subsection{Anchors in Oriented Object Detection}
Oriented object detection is a well-studied research area. The existing CNN-based oriented object detectors can be divided into two categories: anchor boxes \cite{ReDet, R3Det,scrdet, rrpn, zhang2018toward} and dense points \cite{reppoints, wang_2021, fcosr,DARDet, beyond}. A classical solution for the AOOD task is to set rotated anchors \cite{rrpn, liu2016ship}, such as rotated RPN \cite{rrpn}, in which the anchors with different angles, scales, and aspect ratios are placed on each location. These densely rotated anchors lead to extensive computations and memory costs. To address this issue, RoI-Trans \cite{roitrans} models the geometry transformation and solves the problem of misalignment between Region of Interests (RoIs) and objects. Oriented R-CNN \cite{orientedrcnn} designs an oriented RPN to generate oriented proposals directly. Some methods are devoted to improving object representations. Yang et al. \cite{csl} propose CSL to address the boundary problem by transforming angular prediction from a regression problem to a classification task. Gliding Vertex \cite{gliding} glides the vertex of the horizontal bounding box on each corresponding side to accurately describe a multi-oriented object instead of directly regressing the four vertices. These well-designed methods have shown promising performance. However, it still produces many rotated anchors and redundant detection boxes. 

Meanwhile, the keypoint-based AOOD methods have attracted extensive academic attention. These approaches generate the oriented bounding boxes by a set of keypoints belonging to the objects. DARDet \cite{DARDet} proposes a dense anchor-free rotated object detector and an alignment convolution module to extract aligned features. FCOSR \cite{fcosr} develops an ellipse center sampling method for oriented bounding boxes to define the sampling region. CFA \cite{beyond} presents the convex-hull representation to learn the irregular shapes and layouts, which intend to alleviate the feature aliasing.  Generally, these methods require excessive modifications to horizontal anchor-free detectors, which are prone to feature misalignment problems.

As the hand-craft anchor boxes need to be carefully tuned to achieve good performance, while our method tends not to use the anchor boxes. Unlike the above methods, we directly predict the absolute position of each object in the image and remove the need for manual preset boxes and complicated hand-designed components.

\subsection{Label Assignment Strategy for Oriented Object Detection}
The label assignment is a core issue that mainly seeks to define positive/negative training samples independently for each ground-truth object. Anchor-based detectors \cite{faster,ReDet} usually adopt IoU as the assigning criterion. For instance, RPN in Faster R-CNN uses 0.7 and 0.3 as the positive and negative thresholds, respectively. This strategy introduces many hyperparameters that depend on the datasets. It means that one needs to spend much effort adjusting the hyperparameters when the dataset is changed. While the anchor-free detectors directly assign anchor points around the center of objects as positive samples or view each object as a single or a set of keypoints. GGHL \cite{GGHL} proposes a Gaussian OLA strategy to reflect the shape and direction of the object and refine the positive candidate locations. CFA \cite{beyond} categorizes  convex-hulls into positives or negatives according to the CIoU between the convex-hulls and the ground-truth boxes. The remarkable property of non-end-to-end detectors is a one-to-many positive sample assignment. During the training stage, for a ground-truth box, any samples whose confidence threshold is higher than the preset threshold are assigned as the positive samples. It always results in multiple samples in the feature maps being selected as positive samples. As a result, these detectors produce redundant predictions in the inference stage. 

On the contrary, transformer-based detectors apply one-to-one assignments during the training stage \cite{makes}. Our method follows this assignment strategy. For one ground-truth box, only one sample with the minimum matching cost is assigned as the positive sample, and the others are all negative samples. The positive sample is usually selected by bipartite matching to avoid sample conflict. In order to apply the bipartite matching loss in AOOD, we introduce a rotation-aware matching loss to ensure that the entire label assignment process is one-to-one.

\subsection{Transformer Network and its Application}
The transformer is firstly proposed for sequence transduction in \cite{transformer}. The core mechanism of transformer is self-attention which makes it particularly suitable for long-range modeling information contained in all the input tokens. Recently, Carion et al. \cite{detr} present the DETR, which is the first method with an end-to-end optimization objective for set prediction. The series of related works \cite{deformable, anchordetr, conditional, swin, dabdetr} prove that transformers could achieve state-of-the-art performance in image classification and detection. Deformable DETR \cite{deformable} is proposed to combine with sampling deformable points of value to the query and uses multiple level features to solve the slowly converging speed of the transformer detector. In Anchor DETR \cite{anchordetr}, the object queries are based on the anchor points, while Conditional DETR \cite{conditional} encodes the reference point as the query position embedding. 

Transformers are well suited for operating on the points since they are naturally permutation invariant and can enable adaptive receptive fields for oriented objects. We have been inspired to explore the encoder-decoder paradigm for the AOOD task. The self-attention is effective for global-dependency modeling, and it is likely to be valuable for rigid rotating and arbitrary placed objects. Our work is inspired by the recent Deformable DETR and anchor DETR for object detection. Different from them, the proposed AO2-DETR is an arbitrary-oriented end-to-end transformer-based detector, which can be trained from scratch and has significant design differences such as oriented proposal and adaptive refinement module. Overall, the proposed novel designs offer more flexibility with broad context modeling and fewer inductive biases for the AOOD task.

\begin{figure*}[!t] 	
  \centering 	
  \includegraphics[width=\textwidth,height=\textheight,keepaspectratio]{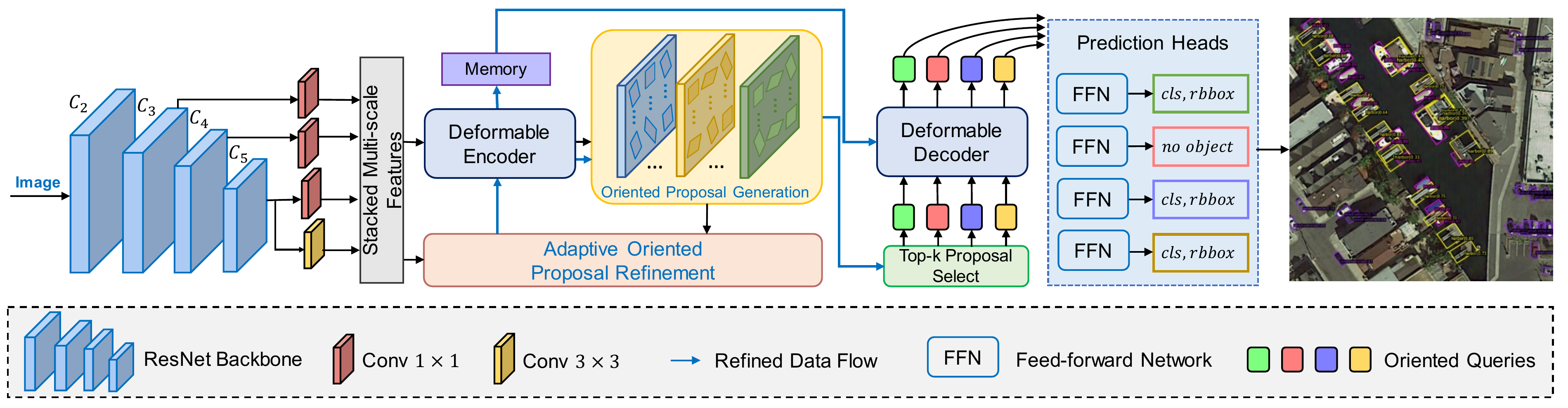}  	
  \caption{Illustration of our proposed framework. AO2-DETR adapts the standard Deformable DETR for the AOOD task by introducing: (i) an oriented proposal generation mechanism to generate oriented proposals as object queries, which provides a better positional prior for pooling features. (ii) an adaptive oriented proposal refinement module to adjust the oriented proposals according to the learned context information, and (iii) a rotation-aware set matching loss to ensure the one-to-one matching process in AO2-DETR. The feed-forward network predicts either a detection (class and bounding box) or a ``no object'' class.} 	
  \label{fig:framework} 
\end{figure*}

\section{Method}
\subsection{Overview}
The framework of the proposed method is shown in Fig. \ref{fig:framework}, which is mainly composed of six components: (1) a CNN backbone, (2) a deformable encoder, (3) an oriented proposal generation mechanism to generate oriented region proposals, (4) an adaptive oriented proposal refinement module to reconstruct the feature map and refine the oriented proposals adaptively, (5) a deformable decoder, (6) and a rotation-aware set matching loss to ensure the correct one-to-one matching process.

AO2-DETR takes an image as input and predicts the positions of objects in the form of oriented bounding boxes $(x, y, w, h, \theta)$ (denoted by OpenCV representation). Given an image, a CNN backbone is firstly used to extract a compact multi-scale feature map. The image features from the CNN backbone are passed through the transformer encoder, together with spatial position encoding that are added to queries and keys at every multi-scale deformable attention module.

Then, the oriented proposal generation mechanism receives the encoder memory to generate oriented region proposals. With these oriented region proposals, we can mitigate the problem of misalignment and cluttered features, thus providing a better positional prior for pooling features to modulate the cross-attention. To extract rotation-invariant region features and eliminate the misalignment between region features and oriented proposals, an adaptive oriented proposal refinement module is proposed to reconstruct feature map and refine the initial oriented proposals. The blue line in Fig. \ref{fig:framework} denotes the refined data flow. Next, we select the top-k scores refined oriented proposals as object queries, which will be fed into the deformable decoder as object queries. 

Consequently, the top-k object queries are transformed into output embeddings by the deformable decoder through multiple multi-head self-attention and multi-scale deformable attention modules. They are then independently decoded into box coordinates and class labels by an FFN, we can obtain the final set of predictions (class and bounding box) or a “no object” class. And the rotation-aware matching loss is proposed to ensure the correct one-to-one matching in the training phase.

\subsection{Backbone and Transformer Encoder}
Starting from the initial image $x_{\mathrm{img}} \in \mathbb{R}^{3 \times H_{0} \times W_{0}}$ (with 3 color channels), a conventional CNN backbone generates multi-scale feature maps $\left\{x^{l}\right\}_{l=1}^{L}(L=4)$ at different resolutions from the output feature maps of stage $C_{3}-C_{5}$.
 The input of encoder is multi-scale feature maps $\left\{x^{l}\right\}_{l=1}^{L-1}(L=4)$ which are extracted from the output feature maps of stages $C_3$ through $C_5$ in ResNet (transformed by a $1 \times 1$ convolution with stride 1) and positional embeddings. The lowest resolution feature map $x^{L}$ is obtained via a $3 \times 3$ convolution with stride 2 on the final $C_{5}$ stage, denoted as $C_{6}$. All the multi-scale feature maps are of $256$ channels.

The transformer encoder is employed to model the discriminative contextual information among all the pixel levels. The key and query elements are pixels from the multi-scale feature maps. For each query pixel, the reference point is itself. To identify the feature level of each query pixel and the positional embedding, we add a scale-level embedding to the feature representation. Unlike the positional embedding with fixed encodings, the scale-level embeddings are randomly initialized and jointly trained with the network. Each encoder layer has a standard architecture consisting of a multi-scale deformable attention module and a fully connected FFN.

\subsection{Oriented Proposals Generation Mechanism}
In the original DETR, object queries in the decoder are irrelevant to the current image. To address this issue, Deformable DETR made some improvements by generating region proposals in the first stage and then providing them into the decoder as object queries in the second stage. However, due to the highly diverse directions of objects in the aerial images, it is intractable to acquire accurate object area by using these horizontal region proposals as object queries. As a result, it usually turns out to be difficult to train a detector for extracting object features and identifying the accurate localization. To address this issue, a novel oriented proposal generation (OPG) mechanism is proposed to produce more accurate oriented proposals by learning the angle of each proposal in addition to the center and size. The generated oriented proposals will be served as object queries in the deformable decoder, which provide a better positional prior for pooling features to modulate the cross-attention, as shown in Fig. \ref{fig:framework}. Specifically, let $i$ index a pixel from feature level $l_{i} \in\{1,2, \ldots, L\}$ with normalized coordinates $\left(p_{ix}, p_{iy}\right) \in[0,1]^{2}$, an initial rotated box is firstly generated as $p_i=(p_{ix}, p_{iy}, p_{iw}, p_{ih}, p_{i\theta})$. Then we use the prediction results $\Delta r_i$ obtained from the image features encoded by the deformable encoder and the initial rotated box $p_i$ to obtain the final oriented proposal $\hat{b}_i$ for each pixel $i$. Here, $\hat{b}_{ij} = \left(x_{ij}, y_{ij}\right) $ denotes the four vertices of $\hat{b}_i$, $j \in\{1,2,3,4\}$, $x_{ij}$ and $y_{ij}$ represents the $x$ and $y$ coordinates of point $\hat{b}_{ij}$, respectively. Then, $\hat{b}_{ij}$ is formulated as:
\begin{equation}
  \begin{aligned} 
    \label{op}
    \hat{b}_{ij}&=(\sigma\left(\sigma^{-1}\left(\phi_{x_j}\left(p_i\right)\right)+ \phi_{x_j}\left(\Delta r_{i}\right)\right), \\&\phantom{=\;\;} \sigma\left(\sigma^{-1}\left(\phi_{y_j}\left(p_i\right)\right)+ \phi_{y_j}\left(\Delta r_{i}\right)\right)),
\end{aligned}
\end{equation}
\begin{equation} 
    \begin{aligned}
    \phi_{x_j}(p_{i})&=p_{ix}+\frac{1}{2}\left(\cos (p_{i\theta}) \times \cos \left(\left\lfloor\frac{j-1}{2}\right\rfloor \pi\right) \times p_{iw}\right.\\&\phantom{=\;\;} \left. - \sin (p_{i\theta}) \times \cos \left(\left\lceil\frac{j+1}{2}\right\rceil \pi\right) \times p_{ih}\right),
    \end{aligned} 
\end{equation}
\begin{equation}
  \begin{aligned}
  \phi_{y_j}(p_{i})&=p_{iy}+\frac{1}{2}\left(\sin (p_{i\theta}) \times \cos \left(\left\lfloor\frac{j-1}{2}\right\rfloor \pi\right) \times p_{iw}\right.\\ &\phantom{=\;\;} \left. +\cos (p_{i\theta}) \times \cos \left(\left\lceil\frac{j+1}{2}\right\rceil \pi\right) \times p_{ih}\right),
  \end{aligned}
\end{equation}
where $p_{iw} $ and $p_{ih}$ are both set as $2^{l_{i}-1}s$, $s=0.05$, $p_{i\theta}$ is set as 0. Here, $\Delta r_{i\{x, y, w, h, \theta\}} \in \mathbb{R}$ are predicted by the bounding box regression branch, $\phi_{x_j}$ and $\phi_{y_j}$ represent the calculation process of the $x$ and $y$ coordinates of the four vertices of oriented boxes, respectively. The calculation process of $\phi_{x_j}\left(\Delta r_{i}\right)$ and $\phi_{y_j}\left(\Delta r_{i}\right)$ are the same as $\phi_{x_j}\left(p_i\right)$ and $\phi_{y_j}\left(p_i\right)$, respectively. And $\sigma$ and $\sigma^{-1}$ denote the sigmoid and the inverse sigmoid function, respectively. The usage of $\sigma$ and $\sigma^{-1}$ is to ensure $\hat{b}_{ij}$ is of normalized coordinates, as $\hat{b}_{ij} \in [0, 1]$. 
\begin{figure}[!t]                        \centering                           
  \includegraphics[width=0.48\textwidth,keepaspectratio]{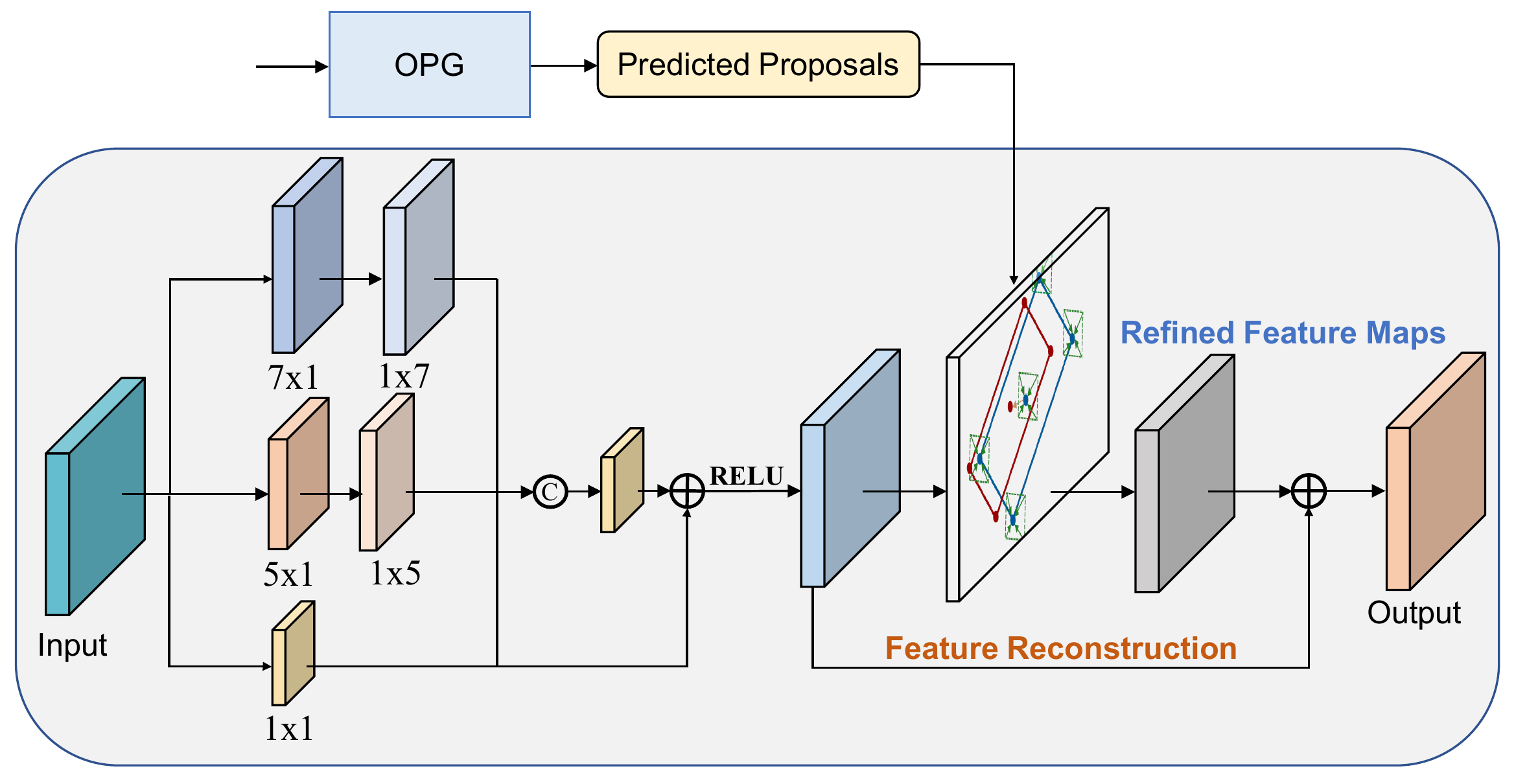}                         
  \caption{Illustration of adaptive oriented proposal refinement (OPR) mechanism. It mainly contains two parts: a three-way convolutional layer and feature alignment module.}                           
  \label{fig:opr}         
\end{figure}

\subsection{Adaptive Oriented Proposal Refinement Module}
Many objects in aerial images are usually distributed with large-scale variations and arbitrary orientations. The convolution features are usually axis-aligned with the fixed receptive field, which will lead to the misalignment between the extracted convolution features and oriented objects and will affect the final detection performance. Therefore, it is crucial to extract rotation-invariant region features and eliminate the misalignment between region features and objects, especially for dense regions. To achieve better detection accuracy, many refined detectors \cite{refinedet, PDC, anchorfree_refine, orientedrcnn, s2anet,R3Det} usually tend to design different feature refinement module. We introduce the oriented proposal refinement (OPR) module to align the convolutional features and oriented proposals.

The structure of the OPR module is shown in Fig. \ref{fig:opr}. The inputs are the multi-scale feature map of the backbone and the initial oriented proposal of the deformable encoder. The output is a refined feature map. To effectively excavate the contextual information, we use the relevant knowledge of the human visual perception system to incorporate more discriminative feature representation by constructing a larger receptive field block. To be specific, the feature map is added by three-way convolution (conv $1\times1$, conv $5\times1$, and conv $7\times1$) to obtain a large kernel receptive fields. Here, $F \in \mathbb{R}^{C \times 1 \times 1}$ represents the feature vector of the point on the feature map. The whole process can be expressed as follows:  
\begin{equation} 	      
  X_{out}=\tau\left(Br_1 \oplus \epsilon(Br_1 \odot Br_2 \odot Br_3 )\right),  
\end{equation}     
where $X_{out}$ represents the output feature, $Br_1$, $Br_2$, and $Br_3$ denote the output of the three branches ``conv $1\times1$", ``conv $1\times5$", and ``conv $1\times7$", respectively. Here, $\oplus$ represents the operation of feature addition, $\odot$ represents the operation of feature concatenation, $\epsilon$ denotes the process of adjusting the number of channels through $1 \times 1$ convolution, $\tau$ is the activation function of ReLU. After the above steps, the OPR module can highlight the relationship between the size and eccentricity of different receptive fields, and force the network to learn discriminative information. Then the new feature map is sent into the feature alignment module.
\begin{figure}[!t]               
  \centering               
  \includegraphics[width=0.48\textwidth,keepaspectratio]{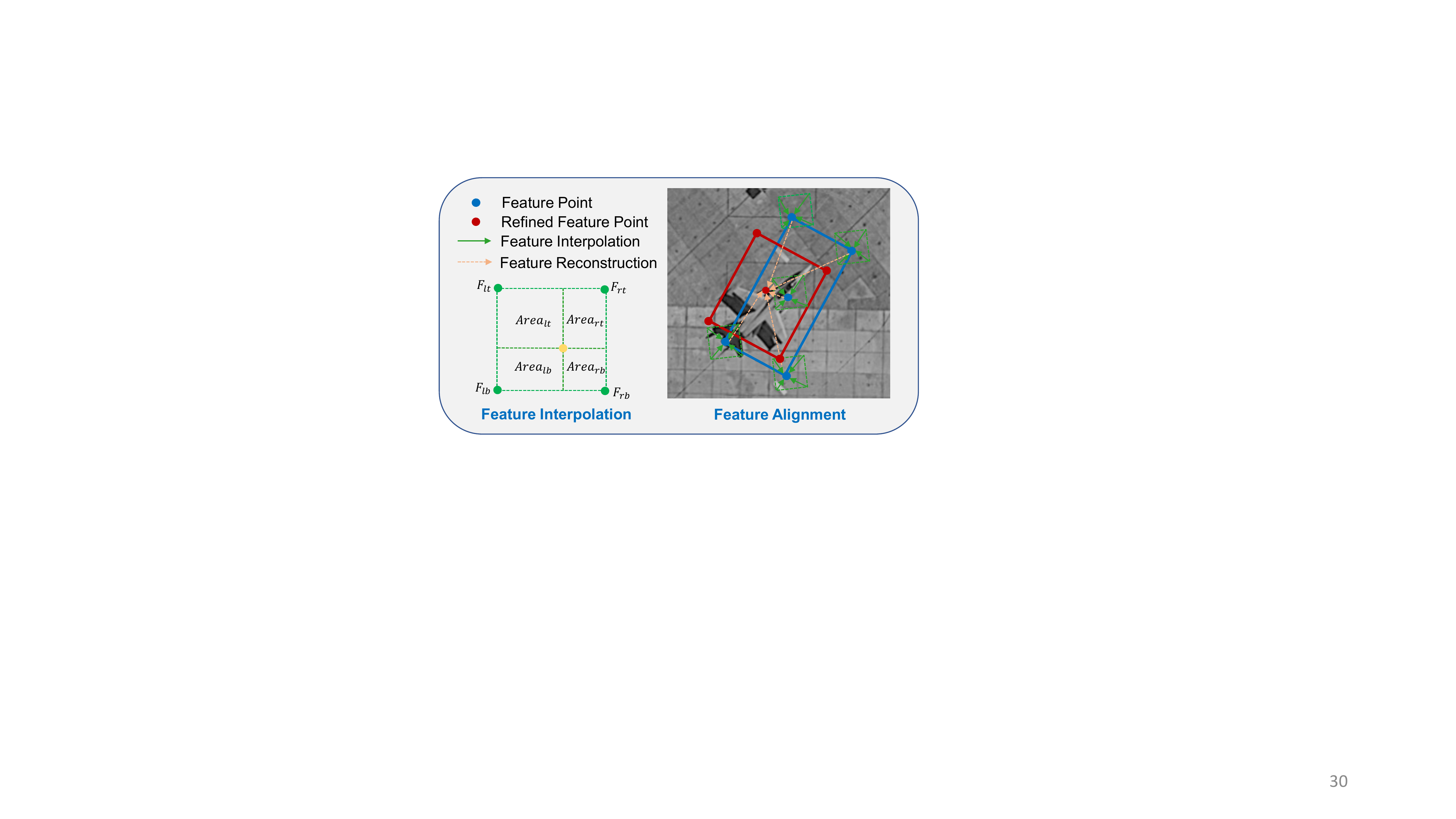}      
  \caption{Illustration of feature alignment module. We aim to re-encode the position information of the initial bounding box (blue rectangle) to the refined bounding box (red rectangle). We adopt the bilinear feature interpolation method as R$^3$Det \cite{R3Det}.}               
  \label{fig:fa}     
\end{figure}

The process of feature alignment is shown in Fig. \ref{fig:fa}. We extend the feature alignment module in \cite{R3Det} into our method. The inputs of this module are the feature map $X_{out}$ and the oriented proposals $\hat{b}_i$ (in Equation \ref{op}). Specifically, we re-encode the position information of the initial oriented proposal (blue rectangle) to the corresponding feature points (red point), thereby reconstructing the entire feature map by a pixel-wise manner to achieve the alignment of the features. We also adopt the bilinear feature interpolation to obtain the feature information corresponding to the initial oriented proposals. The formulation of feature interpolation is as follows:
\begin{equation}
  F=F_{lt} * A_{rb}+F_{rt} * A_{lb}+F_{rb} * A_{lt}+F_{lb} * A_{rt},
\end{equation}
where $A$ denotes the $Area$ in Fig. \ref{fig:fa}, $F$ denotes the final reconstructed feature map, which is added to the original feature map to complete the whole process. The OPR module can capture the arbitrary geometric structure of an oriented proposal and its surrounding context information, which is essential for reducing the misalignment between the predicted oriented proposals and the ground truth one. The reconstructed feature map will be sent to the deformable encoder and OPG module to generate refined oriented proposals. This process is represented by the blue line in Fig. \ref{fig:framework}. Finally, the top-k scores refined oriented proposals will be selected as object queries, where the positional embeddings of object queries are set as positional embeddings of oriented proposal coordinates. Then these object queries are sent into the deformable decoder to output the final set of predictions in parallel.
\begin{figure*}[!t]         
  \centering         
  \includegraphics[width=0.78\textwidth,keepaspectratio]{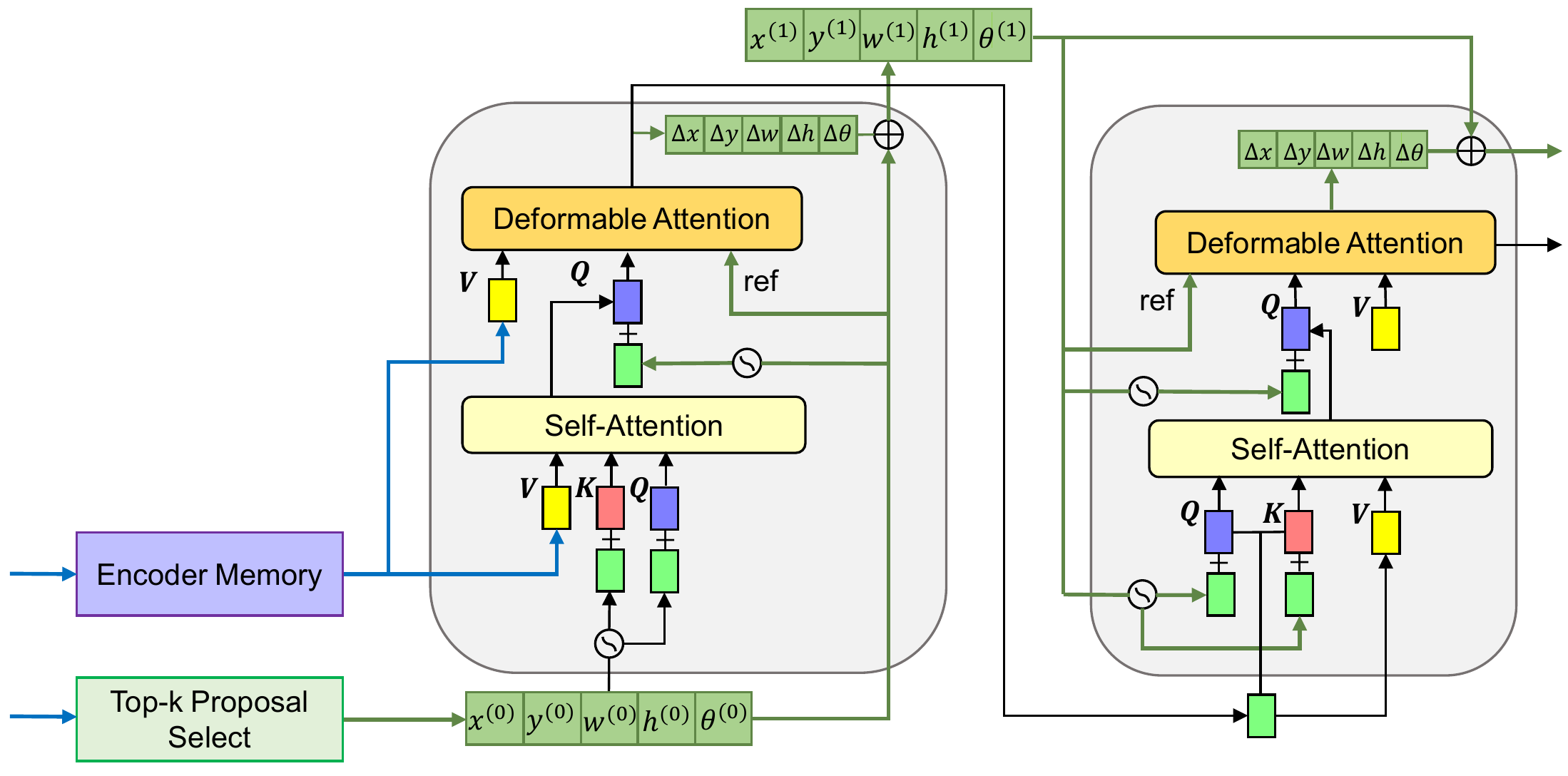}          
  \caption{Illustration of the deformable decoder of the proposed method AO2-DETR. Given the encoder memory and top-k scoring proposals, we perform the deformable cross-attention operation at each reference proposal in the decoder. The object queries are updated layer-by-layer to gradually get close to the ground-truth objects, which provides a better positional prior for pooling features to modulate the cross-attention. The blue line denotes the refined information flow of the OPR module. The green line indicates the information flow of reference proposals.}         
  \label{fig:decoder}   
\end{figure*}

\subsection{Deformable Transformer Decoder}
There are cross-attention and self-attention modules in the decoder. In the cross-attention modules, object queries extract features from the feature maps, where the key elements are of the output feature maps from the OPR module. Following \cite{deformable}, we only replace each cross-attention module to be the multi-scale deformable attention module. The deformable decoder is performed across multi-scale feature maps, encoding richer context over a larger receptive field and enabling the network to learn oriented receptive fields. Thus, we can enable the network to handle object detection with small objects and variable orientations. The output embeddings of the deformable decoder will then be fed into two branches: bounding box regression and classification. The classification $F_{cls}$ is a single layer FFN, while the regression branch $F_{reg}$ is a 3-layer FFN. Consequently, the model globally reasons about all objects and is able to use the whole image as context.

Unlike the original deformable attention module in the decoder of Deformable DETR, the multi-scale deformable attention module in the proposed method attends to a small fixed number of key sampling points around an oriented region proposal, rather than the horizontal proposal. The deformable decoder of our method is shown in Fig. \ref{fig:decoder}. Let $\hat{c}_{q}$ be the oriented region proposal for each query element $q$. Then the multi-scale deformable attention module is applied as: 
\begin{equation}    
  \begin{aligned}   
    \label{eq6}   
    & \operatorname{MSDeformAttn}\left(\boldsymbol{z}_{q}, \hat{\boldsymbol{c}}_{q},\left\{\boldsymbol{x}^{l}\right\}_{l=1}^{L}\right)= \\
    & \sum_{m=1}^{M} \boldsymbol{W}_{m} \left[\sum_{l=1}^{L} \sum_{k=1}^{K} A_{mlqk} \cdot \boldsymbol{W}_{m}^{\prime} \boldsymbol{x}^{l}\left(f_{l}\left(\hat{\boldsymbol{c}}_{q}\right)+\Delta \boldsymbol{p}_{m l q k}\right)\right], 
  \end{aligned}  
\end{equation} 
where $m$ indices the attention head and $K$ is the total number of sampled keys, $l$ indexes the input feature level. $f_{l}\left(\hat{\boldsymbol{c}}_{q}\right)$ rescales the normalized coordinates $\hat{\boldsymbol{c}}_{q}$ to the input feature map of the $l$-th level. $\Delta p_{mlqk}$ and $A_{m l q k}$ denote the sampling offset and attention weight of the $k^{\text {th}}$ sampling point in the $l^{\text {th}}$ feature level and the $m^{\text {th}}$ attention head, respectively. $\boldsymbol{W}_{m}$ and $W_{m}^{\prime}$ are the projection matrices for multi-head attention. The initialization process for multi-scale deformable attention is the same as the Deformable DETR. By using the refined oriented proposals, the learned decoder attention will have a strong correlation with the predicted bounding boxes, which can avoid learning messy information, including background or other objects, especially for densely placed objects, and it can also accelerate the training convergence.  

The detection head predicts the relative offsets, $\hat{b}_q$ is the predicted oriented boxes, the four vertices of $\hat{b}_q$ which is calculated by:
\begin{equation}   
  \begin{aligned}      
    \hat{b}_{qj}&=\{\sigma\left(\phi_{x_j}\left(r_{q}\right)+\phi_{x_j}\left(\sigma^{-1}\left(\hat{c}_{q}\right)\right)\right), \\
    &\phantom{=\;\;} \sigma\left(\phi_{y_j}\left(r_{q}\right)+\phi_{y_j}\left(\sigma^{-1}\left(\hat{c}_{q}\right)\right)\right)\}, 
  \end{aligned} 
\end{equation}
where $j \in\{1,2,3,4\}$, $ r_{q} \in \mathbb{R}$ are predicted by the detection head. Then we express $\hat{b}_q$ as $\{\hat{b}_{qx}, \hat{b}_{qy}, \hat{b}_{qw}, \hat{b}_{qh}, \hat{b}_{q\theta}\}$ by the OpenCV representation of oriented bounding boxes. Crucially there is no down-sampling in spatial resolution but global context modeling at every layer of the transformer decoder, thus offering an entirely new perspective to the oriented object detection. The proposed method simultaneously predicts a set of oriented boxes with no particular ordering.  

In addition, inspired by the iterative refinement developed in Deformable DETR, we also adopt a simple and effective iterative bounding box refinement mechanism to improve detection performance. Here, each decoder layer refines the oriented bounding boxes according to the predictions from the previous layer. Suppose there are $D$ number of decoder layers (e.g., $D$ = 6), given a normalized oriented box $\hat{b}_{q}^{d-1}$ predicted by the $(d-1)^{th}$ decoder layer, the $d^{th}$ decoder layer refines the box as:
\begin{equation}   
  \begin{aligned}      
    \hat{b}_{qj}^{d}&=\{\sigma\left(\phi_{x_j}\left(r_{q}^{d}\right)+\phi_{x_j}\left(\sigma^{-1}\left(\hat{b}_{q}^{d-1}\right)\right)\right), \\
    &\phantom{=\;\;} \sigma\left(\phi_{y_j}\left(r_{q}^{d}\right)+\phi_{y_j}\left(\sigma^{-1}\left(\hat{b}_{q}^{d-1}\right)\right)\right)\}, 
  \end{aligned} 
\end{equation}
where $d \in \{1, 2, ..., D\}$, $\hat{b}_{qj}^{d}$ is the $j^{th}$ vertex coordinates of $\hat{b}_{q}$ at the d-th decoder layer, $r_{q}^{d} \in \mathbb{R} $ are predicted at the d-th decoder layer. Prediction heads for different decoder layers do not share parameters. In the iterative bounding box refinement module, for the $d^{th}$ decoder layer, we sample key elements respective to the box $\hat{b}_{q}^{d-1}$ predicted from the $(d-1)^{th}$ decoder layer. For Equation \ref{eq6} in the cross-attention module of the $d^{th}$ decoder layer, $\hat{b}_{q\{x,y,w,h, \theta\}}^{d-1}$ serves as the new reference oriented proposal.

\subsection{Set Matching and Loss Function}
AO2-DETR infers a fixed-sized sequence of $N$ predictions. One of the main challenges is to score oriented predicted objects with respect to the ground truth. To obtain the box predictions, we apply a 3-layer FFN with ReLU activation function and a linear projection layer to the output embeddings of the deformable decoder. Let $\hat{y} = \left\{\hat{y}_{i}\right\}_{i=1}^{N}$ denote the predicted oriented boxes, and $y$ the ground truth set of objects. Assuming $N$ is larger than the number of objects in the image, we consider $y$ also as a set of size $N$ padded with $\emptyset$ (no object). In order to find a bipartite graph matching between these two sets, we search for a permutation of $N$ elements $\sigma \in O_{n}$ with the lowest cost:
\begin{equation}
  \hat{\sigma}=\underset{\sigma \in O_{n}}{\arg \min } \sum_{i}^{N} \mathcal{L}_{\operatorname{match}}\left(y_{i}, \hat{y}_{\sigma(i)}\right),
\end{equation}
where $\mathcal{L}_{\text {match }}\left(y_{i}, \hat{y}_{\sigma(i)}\right)$ is a pair-wise matching cost between ground truth $y_i$ and a prediction with index $\sigma (i)$. The optimal assignment can be computed efficiently by the Hungarian algorithm \cite{hungarian}.

The matching loss takes the class predictions and the similarity of predicted and ground truth boxes into account. Each element $i$ of the ground truth set can be seen as $y_{i}=\left(c_{i}, b_{i}\right)$ where $c_i$ is the target class label (which may be $\emptyset$) and $b_{i} \in [0, 1] ^5$ is a vector that defines ground truth box center coordinates and its height, width relative to the image size and angle. The long side is height, and the angle range is $[0, pi / 2]$. For the predictions with index $\sigma (i)$, we denote the probability of class $c_i$ as $\hat{p}_{\sigma(i)}\left(c_{i}\right)$ and the predicted oriented bounding box as $\hat{b}_{\sigma(i)}$.

To ensure the correct match between the predicted oriented boxes and ground truth, we add a rotation-aware set matching loss $\mathcal{L}_{\text {riou}}$ in the one-to-one matching process. With the above notation, we define the $\mathcal{L}_{\text {match }}$ as follows:
\begin{equation}
  \begin{aligned}
  \mathcal{L}_{\text {match}}(y_{i}, \hat{y}_{\sigma(i)}) = & \lambda_{\text {cls}} \cdot -\log \hat{p}_{\sigma(i)}\left(c_{i}\right)+ \\
  &\lambda_{\text {L1}} \cdot \mathcal{L}_{\text {box}}\left(b_{i}, \hat{b}_{\sigma}(i)\right) 
  + \\
  &\lambda_{\text {riou}} \cdot \mathcal{L}_{\text {riou}}\left(b_{i}, \hat{b}_{\sigma}(i)\right),
\end{aligned}
\end{equation}
where $c_i \neq \emptyset$. The second and third parts of the matching cost are used to score the bounding boxes. For $\mathcal{L}_{\text {box}}$, we use a linear combination of the SmoothL1 Loss. Here, $\lambda_{\text {cls}}$ and $\lambda_{\text {L1}}$ are the weights of Focal and SmoothL1 set matching loss, respectively.
For the rotation-aware loss $\mathcal{L}_{riou}$, we just simply extended the rotated iou loss \cite{riou} into the Hungarian matching loss. We first compute the coordinates of four vertices $b_{i}=\{(x_1, y_1), (x_2, y_2), (x_3, y_3), (x_4, y_4)\}$, $\hat{b}_{\sigma(i)}=\{(x^{\prime}_1, y^{\prime}_1), (x^{\prime}_2, y^{\prime}_2), (x^{\prime}_3, y^{\prime}_3), (x^{\prime}_4, y^{\prime}_4)\}$, the computation process is formulated as:
\begin{equation}
  \begin{aligned}
    &\text {Area}_{b_{i}}=\mathbf{a} \times \mathbf{b},\\
    &\text{Area}_{\hat{b}_{\sigma(i)}}=\mathbf{a}^{\prime} \times \mathbf{b}^{\prime},\\
    &\mathbf{a}=\sqrt{\left(x_{2}-x_{1}\right)^{2}+\left(y_{2}-y_{1}\right)^{2}},\\
    &\mathbf{b}=\sqrt{\left(x_{2}-x_{3}\right)^{2}+\left(y_{2}-y_{3}\right)^{2}},\\
    &\mathbf{a}^{\prime}=\sqrt{\left(x_{2}^{\prime}-x_{1}^{\prime}\right)^{2}+\left(y_{2}^{\prime}-y_{1}^{\prime}\right)^{2}},\\
    &\mathbf{b}^{\prime}=\sqrt{\left(x_{2}^{\prime}-x_{3}^{\prime}\right)^{2}+\left(y_{2}^{\prime}-y_{3}^{\prime}\right)^{2}},
  \end{aligned}
\end{equation}
then determine the vertices of overlap area if they have, and sort these polygon vertices in anticlockwise order to compute the intersection area $\text { Area }_{\text {overlap }}=\max \left(x_{2}, x_{2}^{\prime}\right)-\min \left(x_{1}, x_{1}^{\prime}\right) \times\left(\max \left(y_{1}, y_{1}^{\prime}\right)-\min \left(y_{2}, y_{2}^{\prime}\right)\right)$, the $\mathcal{L}_{riou}$ is computed as:
\begin{equation}
  \begin{aligned}
    &\text {IoU}=\frac{\text { Area }_{\text {overlap }}}{\text { Area }_{b_{i}}+\text { Area }_{\hat{b}_{\sigma(i)}}-\text { Area }_{\text {overlap }}}, \\
    &L_{riou}= 1 - \text { IoU },
  \end{aligned}
\end{equation}
where $\lambda_{\text {riou}}$ is the weight rotated iou set matching loss. These losses are normalized by the number of objects inside the batch, which helps our model to avoid complex post-processing steps.

\textbf{Loss Function}. L1 loss is used as the regression loss. Rotated IoU loss \cite{riou} is used for the IoU loss computation of two rotated 2D boxes and Focal loss for the classification loss. 

\begin{table*}
  \centering
  \caption{Comparisons with state-of-the-art methods on DOTA-v1.0 OBB Task. $*$ indicates multi-scale training and testing. The results with red and blue colors indicate the best and second-best results of each column, respectively.}
  \label{tab1}
  \begin{adjustbox}{max width=\textwidth}
  \begin{tabular}{l|l|lllllllllllllll|l}
  \hline
  Method & Backbone & PL & BD & BR & GTF & SV & LV & SH & TC & BC & ST & SBF & RA & HA & SP & HC & mAP \\ \hline
  \textbf{Two-stage:}&&&&&&&&&&&&&&&&&\\ \hline
  FR-O \cite{dota}  &resnet101&79.09&69.12&17.17&63.49&34.20&37.16&36.20&89.19&69.60& 58.96& 49.40& 52.52& 46.69&44.80&  46.30&52.93\\
  RoI-Transformer$^*$ \cite{roitrans} & resnet101  &  88.64  &  78.52  &  43.44  &75.92&68.81& 73.68& 83.59&90.74&77.27 & 81.46&58.39&53.54&62.83&58.93&47.67&69.56\\
  SCRDet\cite{scrdet}$^*$ & resnet101&\textcolor{blue}{89.98}&  80.65  &52.09&68.36&68.36&60.32& 72.41&90.85 &\textcolor{red}{87.94}&86.86&65.02&66.68&66.25&68.24&65.21&72.61  \\
  CSL$^*$ \cite{csl}&resnet152&\textcolor{red}{90.25}&\textcolor{blue}{85.53}&54.64&75.31&70.44&73.51&77.62&90.84&86.15&86.69&69.60&68.04&73.83&71.10&68.93&76.17\\
  Gliding Vertex$^*$ \cite{gliding}&resnet101&89.64&85.00&52.26 &77.34&73.01&73.14&86.82&90.74&79.02&86.81&59.55&\textcolor{red}{70.91}&72.94&70.86&57.32&75.02\\
  Oriented R-CNN \cite{orientedrcnn}&resnet50&89.46 & 82.12 & 54.78 & 70.86 & 78.93 & 83.00 & 88.20 & \textcolor{red}{90.90} & 87.50 & 84.68 & 63.97 & 67.69 & 74.94 & 68.84 & 52.28 & 75.87\\
  Oriented R-CNN$^*$ \cite{orientedrcnn}&resnet50&89.84 & 85.43 & \textcolor{red}{61.09} & \textcolor{blue}{79.82} & 79.71 & \textcolor{red}{85.35} & \textcolor{red}{88.82} & 90.88 & 86.68 & \textcolor{blue}{87.73} & \textcolor{red}{72.21} & \textcolor{blue}{70.80} & \textcolor{red}{82.42} & 78.18 & \textcolor{blue}{74.11} & \textcolor{red}{80.87}\\
  ReDet \cite{ReDet} &ReR50-ReFPN&88.79&82.64&53.97&74.00  &78.13&84.06&88.04&\textcolor{blue}{90.89}&\textcolor{blue}{87.78}&85.75&61.76&60.39&75.96&68.07&63.59&76.25 \\ \hline
 \textbf{Single-stage:}&&&&&&&&&&&&&&&&&\\ \hline
  $S^2$A-Net \cite{s2anet}&resnet50&89.11&82.84&48.37&71.11&78.11&78.39&87.25&90.83&84.90&85.64&60.36&62.60&65.26&69.13&57.94&74.12\\
  R$^3$Det \cite{R3Det}&resnet50&89.80&83.77&48.11&66.77&78.76&83.27&87.84&90.82&85.38&85.51&65.67&62.68&67.53&78.56&72.62&76.47\\
  KLD$^*$ \cite{kld}&resnet50&88.91 & 85.23 & 53.64 & \textcolor{red}{81.23} & 78.20 & 76.99 & 84.58 & 89.50 & 86.84 & 86.38 & \textcolor{blue}{71.69} & 68.06 & 75.95 & 72.23 & \textcolor{red}{75.42} & 78.32\\
  DAL\cite{dal}&resnet101&88.68&76.55&45.08&66.80&67.00&76.76&79.74&90.84&79.54&78.45&57.71&62.27&69.05&73.14&60.11 &71.44\\ \hline
  \textbf{Anchor-free:}&&&&&&&&&&&&&&&&&\\ \hline
  IE-Net \cite{ienet} &resnet101&80.20&64.54&39.82&32.07&49.71&65.01&52.58&81.45&44.66&78.51&46.54&56.73&64.40&64.24 &36.75&57.14\\
  Rotated RepPoints \cite{reppoints} & resnet50&83.36&63.71&36.27&51.58&71.06&50.35&72.42&90.10&70.22&81.98&47.46&59.50&50.65&55.51&3.07&59.15\\
  DRN$^*$ \cite{drn} &hourglass104&89.45&83.16&48.98&62.24&70.63&74.25&83.99&90.73&84.60&85.35&55.76&60.79&71.56&68.82&63.92&72.95\\
  DARDet \cite{DARDet}&resnet50&88.89&84.31&55.32&75.49&80.33&81.69&88.24&90.88&83.62&87.46&59.85&65.60&76.86&80.46&65.17&77.61 \\
  DARDet$^*$ \cite{DARDet}&resnet50&89.08 & 84.30 & 56.64 & 77.83 & \textcolor{blue}{81.10} & 83.39 & 88.46 &90.88 & 85.44 & 87.56 & 62.77 & 66.23 & 77.97 & \textcolor{blue}{82.03} & 67.40 & 78.74 \\
  DAFNe$^*$ \cite{DAFNe}&resnet101&89.40&\textcolor{red}{86.27}&53.70&60.51&\textcolor{red}{82.04}&81.17&\textcolor{blue}{88.66}&90.37&83.81&87.27&53.93&69.38&75.61&81.26&70.86&76.95\\
  Oriented RepPoints \cite{oriented_reppoints}&resnet50&87.02&83.17&54.13&71.16&80.18&78.40&87.28&\textcolor{red}{90.90}&85.97&86.25&59.90&70.49 &73.53&72.27&58.97&75.97\\ 
  SASM \cite{sasm} &resnet50&86.42&78.97&52.47&69.84&77.30&75.99&86.72&\textcolor{blue}{90.89}&82.63&85.66&60.13&68.25&73.98&72.22&62.37&74.92\\
  SASM$^*$ \cite{sasm} &resnext 101 &88.41 & 83.32 & 54.00 & 74.34 &80.87& \textcolor{blue}{84.10} & 88.04 & 90.74 & 82.85 & 86.26 & 63.96 & 66.78 & 78.40 & 73.84 & 61.97 & 77.19\\
  CFA\cite{beyond}&resnet101&89.26&81.72&51.81&67.17&79.99&78.25&84.46&90.77&83.40&85.54&54.86&67.75&73.04&70.24&64.96&75.05\\
  CFA\cite{beyond}&resnet152&89.08&83.20&54.37&66.87&81.23& 80.96&87.17&90.21&84.32&86.09&52.34&69.94&75.52&80.76&67.96&76.67\\
  \hline
  \textbf{Ours:}&&&&&&&&&&&&&&&&&\\ \hline
  AO2-DETR & resnet50 &89.27&84.97&56.67&74.89&78.87&82.73&87.35&90.50&84.68&85.41&61.97&69.96&74.68&72.39 &71.62&77.73\\
  AO2-DETR$^*$&resnet50&89.95&84.52&\textcolor{blue}{56.90}&74.83&80.86&83.47&88.47&90.87&86.12&\textcolor{red}{88.55}&63.21&65.09&\textcolor{blue}{79.09}&\textcolor{red}{82.88}&73.46&\textcolor{blue}{79.22}\\ 
  \hline
\end{tabular}
\end{adjustbox}
\end{table*}

\begin{table*}
  \centering
  \caption{Performance comparisons on DOTA-v1.5 test set. $*$ indicates multi-scale training and testing. The results with red and blue colors indicate the best and second-best results of each column, respectively.}
  \label{tab2}
  \begin{adjustbox}{max width=\textwidth}
  \begin{tabular}{l|llllllllllllllll|l} 
  \hline
  Method  & PL  & BD  &BR & GTF & SV & LV   & SH   &TC  & BC   & ST& SBF& RA  & HA  & SP & HC & CC & mAP\\ 
  \hline
  RetinaNet-O \cite{retinanet}    & 71.43 & 77.64& 42.12  & \textcolor{blue}{64.65}  & 44.53& 56.79 & 73.31 & \textcolor{red}{90.84} & 76.02   & 59.96   & 46.95    & 69.24      & 59.65      & 64.52& 48.06   & 0.83 & 59.16 \\
  FR-O \cite{dota}   & 71.89   & 77.64& 44.45 
  & 59.87 & 51.28   & 68.98   & 79.37& \textcolor{blue}{90.78}  
    & \textcolor{blue}{77.38} & 67.50  & 47.75 & 69.72 & 61.22& 65.28 & 60.47   & 1.54    & 62.00\\
  Mask R-CNN-O \cite{mask}   & 76.84                   & 73.51                   & 49.90                   & 57.80                   & 51.31                   & 71.34                   & 79.75                   & 90.46                   & 74.21                   & 66.07                   & 46.21                   & \textcolor{blue}{70.61} & 63.07                   & 64.46                   & 57.81                   & 9.42                    & 62.67                    \\
  HTC-O \cite{htc}          & 77.80                   & 73.67                   & \textcolor{blue}{51.40} & 63.99                   & 51.54                   & 73.31                   & \textcolor{blue}{80.31}                   & 90.48                   & 75.12                   & 67.34                   & 48.51  & \textcolor{red}{70.63}  & \textcolor{blue}{64.84}                   & 64.48 & 55.87                   & 5.15                    & 63.40                    \\  
  \hline
  \textbf{Ours:} &  &  &   &   & & &&&  &  & && &  & &  \\
  AO2-DETR       &\textcolor{blue}{79.55}&\textcolor{blue}{78.14}&42.41&61.23& \textcolor{blue}{55.34}& \textcolor{blue}{74.50}& 79.57& 90.64& 74.76& \textcolor{blue}{77.58} & \textcolor{blue}{53.56}& 66.91& 58.56& \textcolor{blue}{73.11} & \textcolor{blue}{69.64}& \textcolor{blue}{24.71}& \textcolor{blue}{66.26}\\
  AO2-DETR$^*$   & \textcolor{red}{87.13} & \textcolor{red}{85.43} & \textcolor{red}{65.87}  & \textcolor{red}{74.69} & \textcolor{red}{77.46}  & \textcolor{red}{84.13}  & \textcolor{red}{86.19} & 90.23 & \textcolor{red}{81.14} & \textcolor{red}{86.56}  & \textcolor{red}{56.04} & 70.48& \textcolor{red}{75.47} & \textcolor{red}{78.30}  & \textcolor{red}{72.66} & \textcolor{red}{42.62} & \textcolor{red}{75.89}  \\
  \hline
  \end{tabular}
  \end{adjustbox}
  \end{table*}
\begin{table}   
  \centering   
  \caption{Evaluation results on SKU110K-R using the COCO-style metric. The results with red and blue colors indicate the best and second-best results, respectively. } 	
  \label{tab3}   
  \begin{tabular}{llll}   \hline   Method & mAP & AP$_{75}$ & AR$_{300}$ \\ \hline   YoloV3-Rotate \cite{yolov3}&49.1&51.1&58.2\\   CenterNet-4point\cite{centernet} &34.3&19.6&42.2\\
  CenterNet \cite{centernet}&54.7&61.1&62.2\\   
  DRN \cite{drn}&55.9&63.1& 63.3\\   
  CFA \cite{beyond}&\textcolor{blue}{57.0}&\textcolor{blue}{63.5}&\textcolor{blue}{63.9}\\   
  AO2-DETR (Ours)& \textcolor{red}{58.0}& \textcolor{red}{64.2} & \textcolor{red}{64.8} \\ \hline 
  \end{tabular} 
\end{table}

\begin{table*}[!t]   
  \centering   
  \caption{Comparisons with state-of-the-art methods on HRSC2016. $*$ indicates the methods are evaluated under VOC2012 metrics, while other methods are all are evaluated under VOC2007 metrics. The results with red and blue colors indicate the best and second-best results, respectively.}   
  \label{tab4}   
  \begin{adjustbox}{max width=\textwidth}   
    \begin{tabular}{c|ccccccccclll}     \cline{1-10}     Method & BL2   & RC1\&RC2 \cite{RC1} & R2PN \cite{zhang2018toward}  & RRD \cite{liao2018rotation} & RoI Transformer \cite{roitrans} & Gliding Vertex \cite{gliding} & R$^3$Det \cite{R3Det} & DRN \cite{drn}   & CenterMap \cite{centermap} & \multicolumn{1}{c}{} &  &  \\     mAP    & 55.7  & 75.7     & 79.6  & 84.3  & 86.2            & 88.20          & 89.26           & 92.70$^*$ & 92.80$^*$ &                      &  &  \\ \cline{1-10}     
      Method & CSL \cite{csl}   & S$^2$A-Net \cite{s2anet}  & ReDet \cite{ReDet} & MRDet \cite{mrdet} & CFA\cite{beyond} & DAL\cite{dal} & DARDet \cite{DARDet} & DAFNe \cite{DAFNe} & Ours& \multicolumn{1}{c}{} &  &  \\     mAP    & 89.62 & 90.17 / 95.01$^*$   & \textcolor{red}{90.46} / \textcolor{red}{97.63}$^*$ & 89.94 & 93.90           & 89.77          & 90.17 (resnet50) & 81.36 & \textcolor{blue}{88.12} / \textcolor{blue}{97.47}$^*$ (resnet50) &  &  &  \\ \cline{1-10}   
    \end{tabular} 
  \end{adjustbox} 
\end{table*}

\begin{table*}[]   
  \centering   
  \caption{Ablation Studies of proposed modules in AO2-DETR. DOTA-v1.0 is used in this experiment. "Deformable DETR -O" means the standard Deformable DETR by only adding the predicted values of angles. The bold results indicate the best performance.}   
  \label{tab5}   
  \begin{adjustbox}{max width=\textwidth}   
    \begin{tabular}{cccccccccccccccccccccc}   \hline   & OPG & OPR & IBR & Angle Branch & PL & BD & BR & GTF & SV & LV & SH & TC & BC & ST & SBF & RA & HA & SP & HC & mAP \\ \hline   
      Deformable DETR-O \cite{deformable} &&&&&75.87 & 56.79 &17.38 &39.14& 36.31&13.79&19.39&87.65&66.54&48.81&18.99&47.59&29.30&52.73&17.58&41.86\\   
      AO2-DETR &\cmark&&&&89.13&78.23&53.27&69.08&54.60&78.52&79.25&90.42&75.63&77.04&56.69&70.61&65.15 &68.43&63.46&71.30\\   
      AO2-DETR &\cmark&\cmark &&&89.16&82.69&55.14&73.25&78.09&80.56&85.46&89.65&83.41&82.52&60.82 &69.68&69.78&73.88&69.39&76.23\\   
      AO2-DETR &\cmark&\cmark& \cmark&& 89.27&84.97&56.67&74.89&78.87&82.73&87.35&90.50&84.68&85.41&61.97&69.96&74.68&72.39 &71.62&\textbf{77.73}\\ 
      AO2-DETR &\cmark&\cmark&\cmark&\cmark& 83.68 & 63.12 & 36.74 & 51.89 & 71.60 & 50.57 & 72.24 & 90.02 & 70.23 & 81.82 & 47.65 & 59.09 & 50.52 & 55.18 & 03.77 &59.20 \\ 
      \hline 
    \end{tabular} 
  \end{adjustbox} 
\end{table*}

\begin{table*} 
  \centering \caption{Ablation studies of set matching cost. DOTA-v1.0 is used in this experiment. The bold results indicate the best performance. The numbers with blue color indicates the performance gain.} 
  \label{tab6} 
  \begin{tabular}{c|ccccc|c}  \hline & Focal Loss Cost & L1 Loss Cost & Rotated IoU Loss Cost & GWD Loss Cost & KLD Loss Cost & mAP    \\  \hline AO2-DETR &\cmark&\cmark&\xmark&\xmark&\xmark&56.65  \\ 
  AO2-DETR& \cmark&\cmark& \cmark&\xmark&\xmark&\textbf{77.73} (\textcolor{blue}{+21.08})   \\ 
  AO2-DETR& \cmark &\xmark&\cmark&\xmark&\xmark&72.68 (\textcolor{blue}{+16.03})\\ 
  AO2-DETR & \cmark& \cmark &\xmark&\cmark&\xmark&58.13 (\textcolor{blue}{+1.48})   \\ 
  AO2-DETR & \cmark& \cmark&\xmark&\xmark&\cmark& 59.76 (\textcolor{blue}{+3.11})   \\ 
    \hline 
  \end{tabular} 
\end{table*}

\section{Experiments}
\subsection{Datasets}
To demonstrate the effectiveness for the proposed method, we conduct experiments on four oriented datasets, DOTA-v1.0 \cite{dota}, DOTA-v1.5 \cite{dota}, SKU110K-R \cite{drn}, and HRSC2016 \cite{hrsc2016} datasets.

\textbf{DOTA} is one of the largest dataset for oriented object detection with two released versions: DOTA-v1.0 and DOTA-v1.5. \textbf{DOTA-v1.0} contains 2,806 large aerial images, and the image size ranges from around $800\times800$ to $4000\times4000$ and $188,282$ instances among 15 common categories: \emph{Plane (PL), Baseball diamond (BD), Bridge (BR), Ground track field (GTF), Small vehicle (SV), Large vehicle (LV), Ship (SH), Tennis court (TC), Basketball court (BC), Storage tank (ST), Soccer-ball field (SBF), Roundabout (RA), Harbor (HA), Swimming pool (SP), and Helicopter (HC)}. \textbf{DOTA-v1.5} is released with a new category, \emph{Container Crane (CC)}. It contains 402,089 annotated object instances within 16 categories. 

We use both training and validation sets for training, the test set for testing. We crop the images into $1024 \times 1024$ patches with a stride of 824. The random horizontal flipping is adopted to avoid over-fitting during training and no other tricks are utilized. For fair comparisons with other methods, we adopt data augmentation at three scales \{0.5, 1.0, 1.5\} and random rotation from 5 angles $\{30^{\circ}, 60^{\circ}, 90^{\circ}, 120^{\circ}, 150^{\circ}\}$. The performance of the test set is evaluated on the official DOTA evaluation server\footnote{https://captain-whu.github.io/DOTA}.

\textbf{SKU110K-R} is a challenging dataset for commodity detection. It is an extended version of SKU-110K\cite{sku110k}. The images are collected from supermarket stores around the world and include scale variations, viewing angles, lighting conditions, noise levels, and other sources of variability. The original SKU-110K dataset contains 11,762 images in total (8,233 for training, 588 for validation, and 2,941 for testing). The SKU110K-R dataset performs data augmentation by rotating the image 6 different angles $\{-45^{\circ}, -30^{\circ}, -15^{\circ}, 15^{\circ}, 30^{\circ}, 45^{\circ}\}$ on the original dataset. After the augmentation, the number of training, validation, and testing images are 57,533, 4,116, and 20,587, respectively. Each image contains an average of 154 tightly packed objects, up to 718 objects. The image size ranges from $1840\times1840$ to $4320\times4320$. We resize the input image to $800\times800$ and apply random rotation as DOTA dataset. For the SKU110K-R dataset, the results follow standard COCO-style Average Precision (AP) metrics that include $AP_{75}$ (IoU = 0.75), mAP and $AR_{300}$.

\textbf{HRSC2016} contains images of ships at the wharf, which are collected from six famous harbors. It only contains one category ``ship". The image size ranges from $300\times300$ to $1500\times900$. The HRSC2016 dataset contains 1061 images in total (436 for training, 181 for validation, and 444 for testing). We use both training and validation sets for training and the test set for testing. Random horizontal flipping is applied during training. For the detection accuracy on the HRSC2016, we adopt the mean average precision (mAP) as evaluation criteria, which is consistent with PASCAL VOC2007 and VOC2012.

\subsection{Implementation Details}
We implement the proposed method AO2-DETR on MMRotate \cite{mmrotate}. In all experiments, we adopt Deformable DETR \cite{deformable} with ResNet-50 backbone (pre-trained on ImageNet \cite{imagenet}) as the baseline method. Multi-scale feature maps are extracted from \emph{$conv_3$} to \emph{$conv_5$} of ResNet-50. The transformer encoder-decoder follows the same architecture as in Deformable DETR. The number of object queries is set to 300. We train the network with AdamW\cite{adamw} for 50 epochs. In the first 40 epochs, the learning rate is $1e-4$ and then $1e-5$ for another 10 epochs. The momentum and weight decay are 0.9 and 0.0001, respectively. Our method is trained on 3 GeForce RTX 3090 GPUs with a total batch size of 4 for training and a single 3090 GPU for inference. The loss weight $\lambda_{\text {cls}}$, $\lambda_{\text {L1}}$ and $\lambda_{\text {riou}}$ are set as 5, 5, and 8, respectively. In the inference stage, we follow the same scale setting as training. No post-processing is needed for associating objects.

\subsection{State-of-the-art Comparison}
We compare AO2-DETR against some state-of-the-art methods in oriented datasets, the results are shown in Table \ref{tab1}, \ref{tab2}, \ref{tab3}, and \ref{tab4}.
\begin{figure*}[t]         
  \centering         
  \includegraphics[width=\textwidth,keepaspectratio]{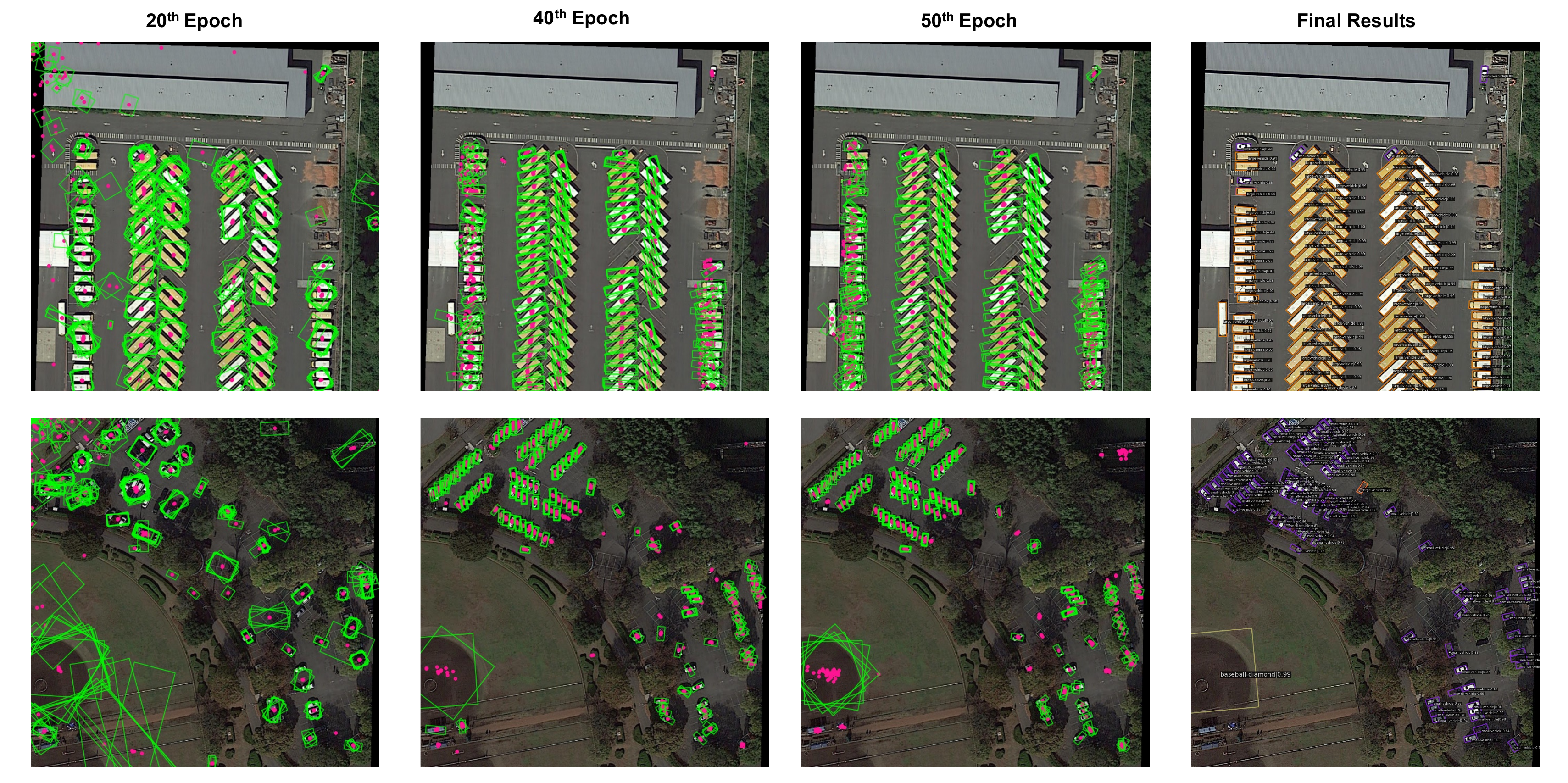}          
  \caption{Visualization of oriented proposal generation (OPG) mechanism. The learned oriented proposals are generated by OPG on the DOTA dataset. The top-300 proposals (green rectangle) per image are displayed. Each sampling point is marked as a pink filled circle. We can see that the sampling points are concentrated inside the object, which is the main focus of deformable attention module.}         
  \label{fig:proposals}   
\end{figure*}

\begin{figure}[!t]   
  \centering   
  \includegraphics[width=0.40\textwidth,keepaspectratio]{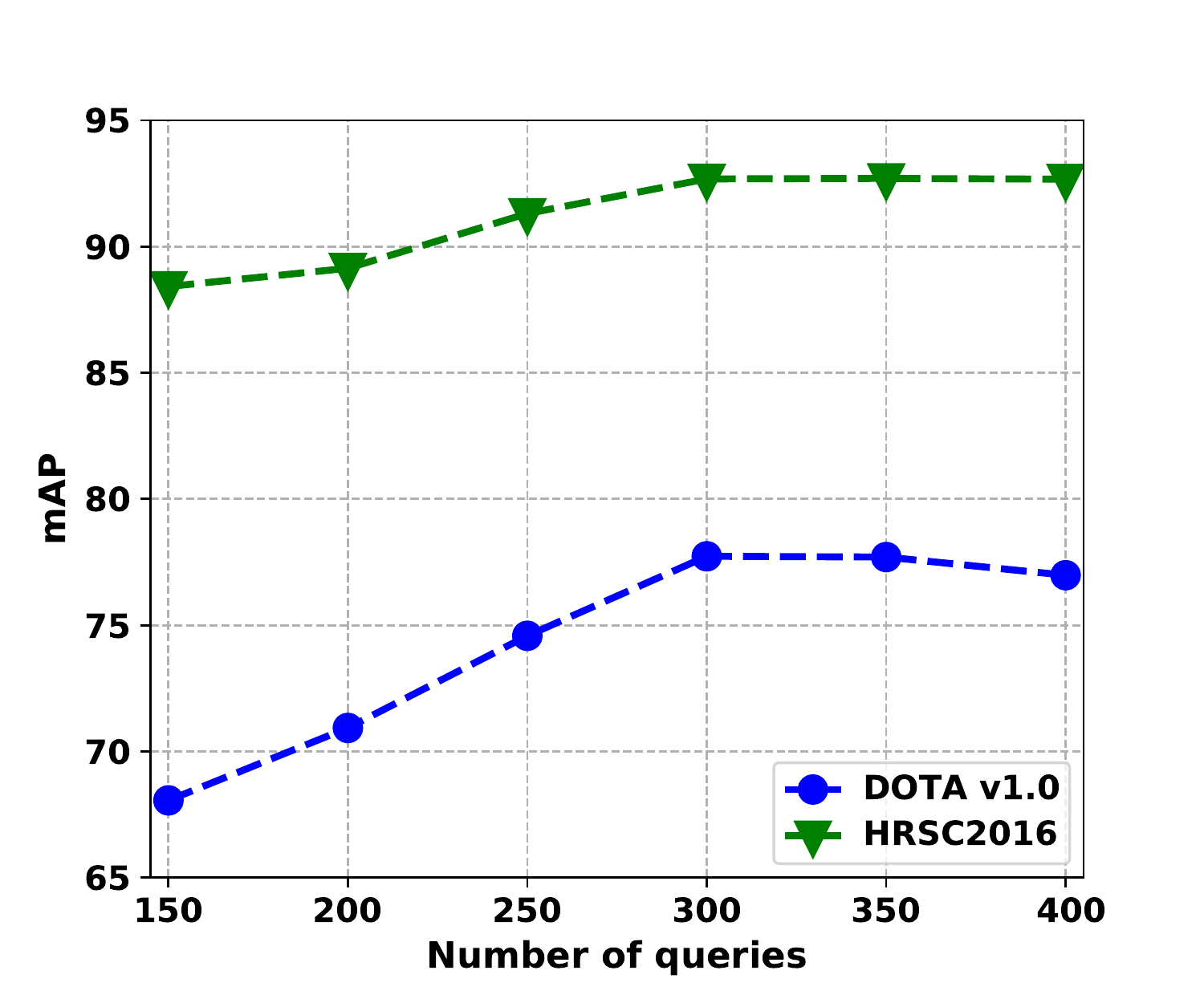}    
  \caption{The ablation results from adapting the number of queries on DOTA-v1.0 and HRSC2016 datasets. We compare the mAP with varying numbers of queries from 150 to 400.}   
  \label{fig:queries} 
\end{figure}

\textbf{Results on DOTA-v1.0}. Table \ref{tab1} shows a comparison of our AO2-DETR with the recently state-of-the-art detectors on the DOTA-v1.0 dataset with respect to oriented bounding box detection. For the accuracy measured by mAP, we achieve 77.73\% mAP with single-scale data and 79.22\% mAP with multi-scale data. Using the same backbone of ResNet50, AO2-DETR achieves the best result among  single-stage and anchor-free methods (e.g., R$^3$Det \cite{R3Det}, KLD \cite{kld}, CFA \cite{beyond}, SASM \cite{sasm}, and DARDet \cite{DARDet}) using a single model without bells and whistles. Specifically, AO2-DETR outperforms R$^3$Det by 1.26\% (77.73\% VS 76.47\%), KLD by 0.9\% (79.22\% VS 78.32\%), CFA by 4.17\% (79.22\% vs 75.05\%), SASM by 2.03\% (79.22\% vs 77.19\%), and DARDet by 0.48\% (79.22\% vs 78.74\%), which is a large margin.

Compared with the anchor-based methods, our method is better than most two-stage methods, except for the best two-stage method Oriented R-CNN \cite{orientedrcnn}. We achieve the second superior performance compared with all recently two-stage methods. Although our method does not achieve the best performance, the proposed method has some apparent advantages over the anchor-based methods. We argue that the superior performance of Oriented R-CNN comes from the oriented region proposal network and a midpoint offset representation to represent oriented objects. Different from existing methods, we aim to design a conceptually simple and a new paradigm framework for the AOOD task. It does not require hand-designed components, complex pre/post-processing steps and inductive biases.

\begin{figure}[!t]   
  \centering   
  \includegraphics[width=0.40\textwidth, keepaspectratio]{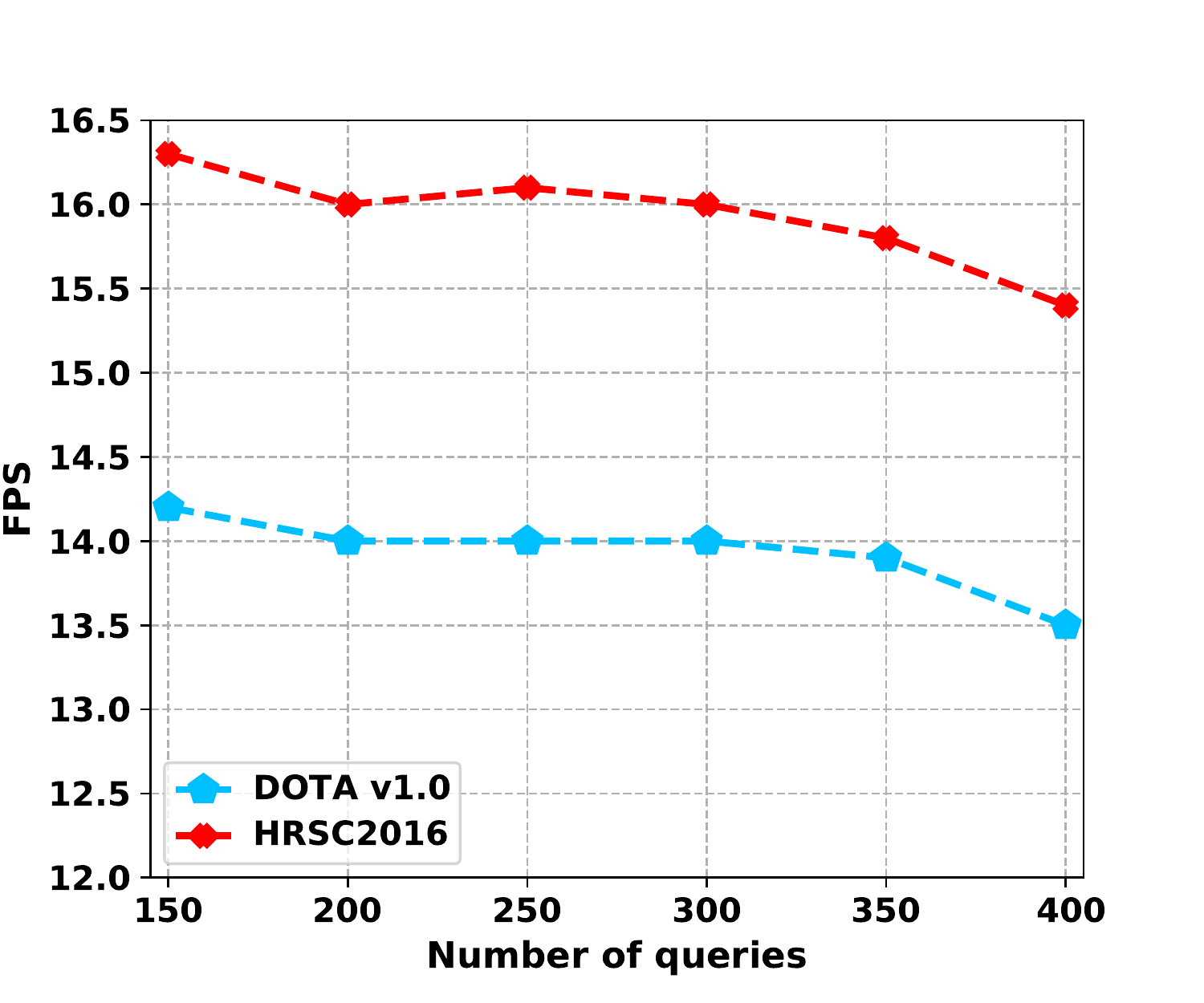}    
  \caption{The ablation results from adapting the number of queries on DOTA-v1.0 and HRSC2016 datasets. We compare the FPS with varying numbers of queries from 150 to 400.}   
  \label{fig:fps} 
\end{figure}

\textbf{Results on DOTA-v1.5}. Compared with DOTA-v1.0, DOTA-v1.5 contains more extremely small objects. We summarize the results on DOTA-v1.5 in Table \ref{tab2}. Compared to state-of-the-art methods, AO2-DETR can achieve 66.26\% mAP with single-scale data and 75.89\% mAP with multi-scale data, outperforms RetinaNet OBB \cite{retinanet}, Faster R-CNN OBB \cite{dota}, Mask R-CNN OBB \cite{mask}, and HTC \cite{htc} by a large margin. These experiments validate that an encoder-decoder detection model based on the standard Transformer can also achieve good results in small object detection, on the basis of not using FPN.

\textbf{Results on SKU110K-R.} The comparison results on SKU110K-R are shown in Table \ref{tab3}. AO2-DETR achieves 58.0\% AP and improves the state-of-the-art anchor-free methods by 1.0\% (58.0\% vs. 57.0\%). We also report the results of $AP_{75}$ and $AR_{300}$. Most of the images in this dataset are taken with handheld cameras, the commodity is placed in a messy way, and the angle changes are relatively large. The results on the SKU110K-R dataset show that our method has strong applicability to this scenario and expands the use of the transformer-based method in the field of intelligent supermarkets.

\textbf{Results on HRSC2016.} The HRSC2016 dataset only contains one category ``ship", which some of them have large aspect ratios and various orientations. The results are shown in Table \ref{tab4}. It can be seen that our AO2-DETR achieves state-of-the-art performance consistently, without the use of a more complicated architecture. Specifically, AO2-DETR achieves 92.68\% and 97.87\% under PASCAL VOC 2007 and VOC 2012 metrics, respectively.

\begin{figure*}[!t]      
  \centering      
  \includegraphics[width=0.95\textwidth,keepaspectratio]{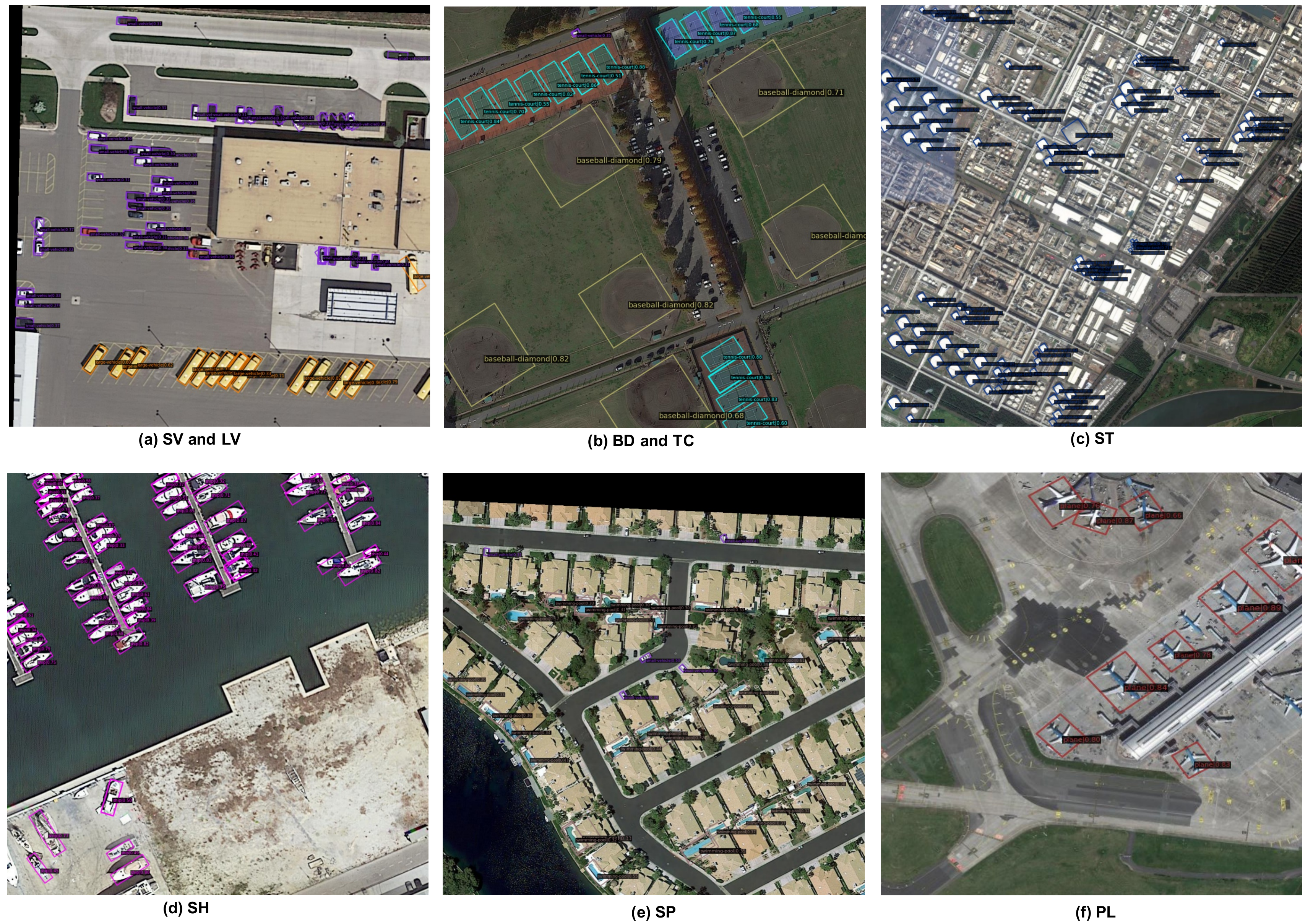}       
  \caption{The qualitative results on DOTA-v1.0 \cite{dota} testing set using AO2-DETR with ResNet50 backbone. DOTA-v1.0 contains 15 common categories, such as large-vehicle, small-vehicle, plane, swimming-pool, ship, tennis-court, etc. The confidence threshold is set to 0.3 when visualizing these results. One color stands for one object class. Best viewed in color and with zoom.}      
  \label{fig:dotav1}  
\end{figure*}

\subsection{Ablation Studies}
In this section, we conduct a series of ablation experiments on the DOTA-v1.0 test set to evaluate the effectiveness of our proposed method. To detect oriented objects, we improve the Deformable DETR by adding the predicted value of the angle and term the detector "Deformable-O". Table \ref{tab5} shows the impact of progressively integrating the proposed components into the baseline framework for the transformer-based AOOD methods.

\textbf{Oriented Proposal Generation Mechanism.}
To explore the contribution of the oriented proposals generation (OPG) mechanism, we derive two settings: with / without oriented proposals generation. The results in Table \ref{tab5} clearly show that the oriented proposals generation is necessary for boosting performance. OPG improves the mAP by 29.44\% (71.30\% vs. 41.86\%). The significant performance improvement indicates that the OPG mechanism can better locate objects and avoid the extracted features from being interfered with by other objects or backgrounds.  

For a better understanding of the role of OPG mechanism, we visualize the adaptation process of sampling points (pink filled circle) and oriented proposals (green rectangle) of the last layer in decoder from 20$^{th}$, 40$^{th}$, and 50$^{th}$ epoch, as shown in Fig. \ref{fig:proposals}. For readability, we combine the sampling points and oriented proposals from feature maps of different resolutions into one image. It can be seen that the sampling points of the 20$^{th}$ epoch are still scattered, but the sampling points of the 40$^{th}$ epoch are concentrated in the center of the objects. At the 50$^{th}$ epoch, the sampling points and generated oriented proposals can well cover the objects. In particular, these green rectangles are only proposals. They will be selected as object queries and sent to the deformable decoder layer to generate the final predictions. The final detection results are displayed in the last column. The visualization results show that the proposed component OPG presents a robust performance in both arbitrary orientation and densely packed scenarios. 

\textbf{Adaptive Oriented Proposal Refinement Module.}
To investigate the importance of the adaptive oriented proposal refinement (OPR) module, we perform a study of models with or without the OPR module. As shown in Table \ref{tab5}, the proposed oriented proposal refinement significantly improves performance by 4.93\% (76.23\% vs. 71.30\%). In addition, we also explore the importance of iterative bounding box refinement. As shown in Table \ref{tab5}, the performance of without iterative refinement module is reduced from 77.73\% to 70.24\%, which shows that the iterative refinement module can enhance the relationship between transformer decoder and better model context information. The both refinement modules provide more precise features for oriented object detection. The proposed OPG and OPR module can effectively mitigate the problems of misalignment and cluttered features.

\begin{figure*}[!t]      
  \centering      
  \includegraphics[width=0.95\textwidth,keepaspectratio]{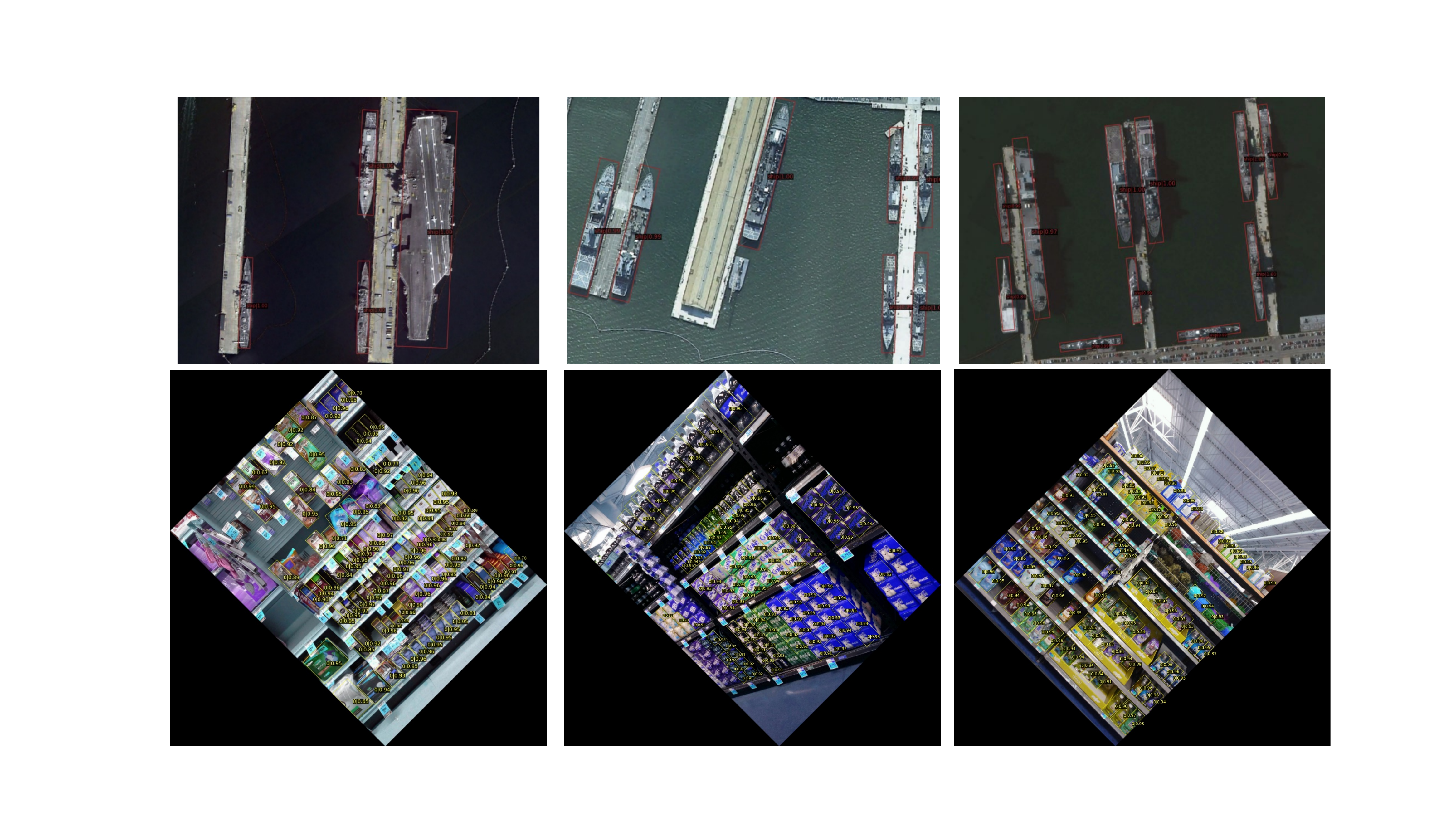}       
  \caption{Detection results on HRSC2016 \cite{hrsc2016} (the first row) and SKU110K-R \cite{drn} (the second row). Best viewed in color and with zoom. }      
  \label{fig:sku}  
\end{figure*}

\textbf{Set Matching Loss.}
For classification and bounding box distance loss, we follow the default settings of the Deformable DETR, i.e., Focal loss and L1 loss. For rotation-aware set matching loss, we have tried different set matching losses and different combinations, such as GWD \cite{gwd}, KLD \cite{kld}, and Rotated IoU \cite{riou}. The experimental results in Table \ref{tab6} prove that Rotated IoU performs best. With the rotation-aware set matching loss, the performance has been greatly improved, from 56.65\% to 77.73\%. We can ensure the correct one-to-one matching process in AO2-DETR by adding the rotation-aware set matching loss, which solves the problem of limited matching. Other existing loss works also can be easily extended to AO2-DETR, but exploring the importance of losses is not the focus of this paper.

\textbf{How about adding an angle prediction branch?} In this experiment, we investigate the impact of the predicted values of oriented bounding boxes. Compared with the direct prediction of five parameters, another option is to add an angle prediction branch to predict angle $\theta$ and maintain the original regression branch to predict $\{x, y, w, h\}$. As shown in Table \ref{tab5}, the performance with an angle prediction branch is reduced from 77.73\% to 59.20\%, which indicates that adding an angle prediction branch does not benefit performance, but complicates the overall structure. It is possible that because of the discontinuity of angles, the separate prediction makes the overall prediction difficult to converge.

\textbf{Number of Queries.}
In this experiment, we compare changing the number of queries at the test stage to different models trained with varying numbers of queries, as shown in Fig. \ref{fig:queries}. As we increase the number of queries, AO2-DETR predicts more bounding boxes, resulting in better performance at the cost of longer running time. Taking the DOTA-v1.0 dataset as an example, we can observe that with the gradual increase in the number of queries, the performance has been significantly improved. When the number of queries is set to 150, the mAP is only 68.06\%, but when the number of queries is increased to 300, the performance can achieve 77.73\% (improved by 9.67\%). This indicates that the number of queries is sufficient to cover the objects well. When we increase the number of queries to 350 and 400, there is a slight decline in performance. We suspect that redundant object queries will interfere with some dense objects, resulting in performance fluctuations.

In addition, we compare the FPS with varying numbers of queries from 150 to 400, as shown in Fig. \ref{fig:fps}. We can see that the number of queries has a smaller impact on FPS, and when the number of queries increases from 150 to 400, the FPS drops from 14.2 to 13.5 for the DOTA-v1.0 dataset, and 16.3 to 15.4 for the HRSC2016 dataset. To achieve a balance of speed and performance, the number of queries is set to 300. For the DOTA-v1.0 dataset, the FPS is 14.0, and for the HRSC2016 dataset, the FPS is 16.0.

With these ablation studies, we conclude that in the AO2-DETR design: oriented proposal generation, adaptive oriented proposal refinement, iterative bounding box refinement, and rotation-aware matching loss all play important roles in the final performance.

\subsection{Qualitative Results}
Fig. \ref{fig:dotav1} and Fig. \ref{fig:sku} show the qualitative results of sample images from the DOTA, HRSC2016, and SKU110K-R datasets. We can notice that the proposed method can deal with various challenges in the AOOD task detection, including multi-oriented objects, small objects, large aspect ratio objects, and densely-packed objects. For example, as shown in Fig. \ref{fig:dotav1}, AO2-DETR can properly detect objects of various sizes and arbitrary orientations within the multi-category classification problem; as shown in Fig. \ref{fig:sku}, AO2-DETR is outstanding in the detection of large aspect ratio objects (the first row) and densely arranged oriented objects (the second row). However, there are still some failure cases, especially when the objects are placed heavily occluded, the sizes of objects are extremely small, and the color of objects is similar to the background. Future transform-based AOOD methods can focus on addressing these difficult cases.

\section{Conclusion}
In this paper, we propose an end-to-end transformer-based detector AO2-DETR for arbitrary-oriented object detection. The proposed AO2-DETR comprises dedicated components to address AOOD challenges, including an oriented proposal generation mechanism, an adaptive oriented proposal refinement module, and a rotation aware set matching loss in order to accurately detect oriented objects in images. The encoder-decoder architecture transforms the oriented proposals (served as object queries) into each corresponding object, which eliminates the need for hand-designed components and complex pre/post-processing. Our approach achieves state-of-the-art performance compared to recently anchor-free and single-stage methods on the oriented datasets (DOTA, HRSC2016 and SKU110K-R datasets). We validate that the transformer can enable adaptive receptive fields for oriented objects, thus it can deal with oriented and irregular placed objects naturally. Furthermore, we hope that this encoder-decoder paradigm will promote future works in oriented object detection.

\textbf{Limitations.} Compared with other CNN-based methods, the main limitation of our method lie in the longer training convergence time. It is widely known that the superior performance of transformers requires relatively larger computation cost. The future work of transformer-based AOOD methods can be devoted to solving these challenges.

\vspace{11pt}

\bibliographystyle{IEEEtran}
\bibliography{reference}
\vspace{-23pt} 
\vspace{11pt}

\vfill

\end{document}